\documentclass[letterpaper]{article} 
\usepackage{times}  
\usepackage{helvet}  
\usepackage{courier}  
\usepackage[hyphens]{url}  
\usepackage{graphicx} 
\usepackage{natbib}  
\usepackage{caption} 
\usepackage{algorithm}
\usepackage{algorithmic}

\usepackage{newfloat}
\usepackage{listings}
\DeclareCaptionStyle{ruled}{labelfont=normalfont,labelsep=colon,strut=off} 
\floatstyle{ruled}
\newfloat{listing}{tb}{lst}{}
\floatname{listing}{Listing}

\title{DisCo-DSO: Coupling Discrete and Continuous Optimization for Efficient Generative Design in Hybrid Spaces}

\author{%
  Jacob F. Pettit,
  Chak Shing Lee,
  Jiachen Yang,
  Alex Ho,\\
  Daniel Faissol,
  Brenden Petersen,
  Mikel Landajuela\thanks{Corresponding author: \texttt{landajuelala1@llnl.gov}}\\
  \\
  Computational Engineering Division\\
  Lawrence Livermore National Laboratory\\
  Livermore, CA, 94550, USA
}

\graphicspath{{./}}
\usepackage{xspace}
\usepackage{multirow}
\usepackage{subfigure}
\usepackage{tikz}
\usepackage{enumitem}
\usepackage{amsmath}

\usepackage{amsmath,amsfonts,bm}

\def\eqref#1{equation~\ref{#1}}

\def\1{\bm{1}}

\def\eps{{\epsilon}}

\DeclareMathAlphabet{\mathsfit}{\encodingdefault}{\sfdefault}{m}{sl}
\SetMathAlphabet{\mathsfit}{bold}{\encodingdefault}{\sfdefault}{bx}{n}

\DeclareMathOperator*{\argmax}{arg\,max}

\newcommand{\ALGNAME}{DisCo-DSO\xspace}
\newcommand{\DT}{decision tree\xspace}
\newcommand{\DTs}{decision trees\xspace}
\newcommand{\defeq}{\overset{\text{\tiny def}}{=}}
\newcommand{\NN}{\text{AR}}
\newcommand{\Lcal}{\mathcal{L}}
\newcommand{\Rbb}{\mathbb{R}}

\usepackage{hyperref} 
\usepackage{natbib} 
\usepackage{cleveref}

\begin{document}

\maketitle

\begin{abstract}
We consider the challenge of
black-box optimization
within hybrid discrete-continuous and variable-length spaces,
a problem that arises in various applications, such as decision tree
learning and symbolic regression. We propose \ALGNAME (Discrete-Continuous Deep Symbolic Optimization), 
a novel approach that uses a generative model to learn a joint distribution
over discrete and continuous design variables to sample new hybrid designs. 
In contrast to standard decoupled approaches, in which the discrete and
continuous variables are optimized separately, our joint optimization
approach uses fewer objective function evaluations, is robust against
non-differentiable objectives, and learns from prior samples to guide
the search, leading to significant improvement in performance and
sample efficiency. Our experiments on a diverse set of optimization tasks
demonstrate that the advantages of \ALGNAME become increasingly evident as the complexity of the problem increases. In particular, we illustrate \ALGNAME's 
superiority over the state-of-the-art methods for interpretable reinforcement
learning with decision trees.

\end{abstract}

\section{Introduction}

\begin{figure}[h!]
    \begin{center}     
    \subfigure[Standard decoupled approach]{
        \centering
        \label{fig:continuous-discrete-decoupled}
        \includegraphics[width=0.51\linewidth]{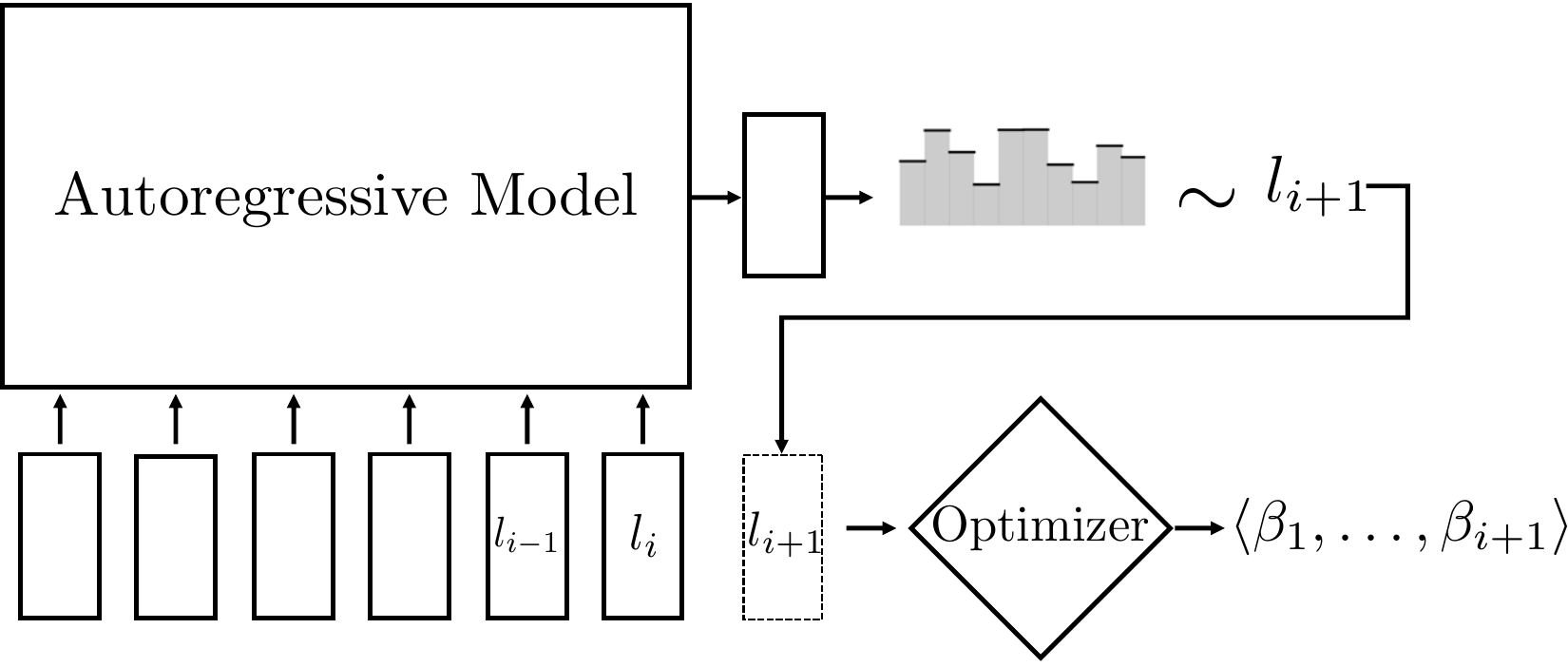}}
    \hfil
    \subfigure[\ALGNAME]{
        \label{fig:continuous-discrete-disco}
        \includegraphics[width=0.44\linewidth]{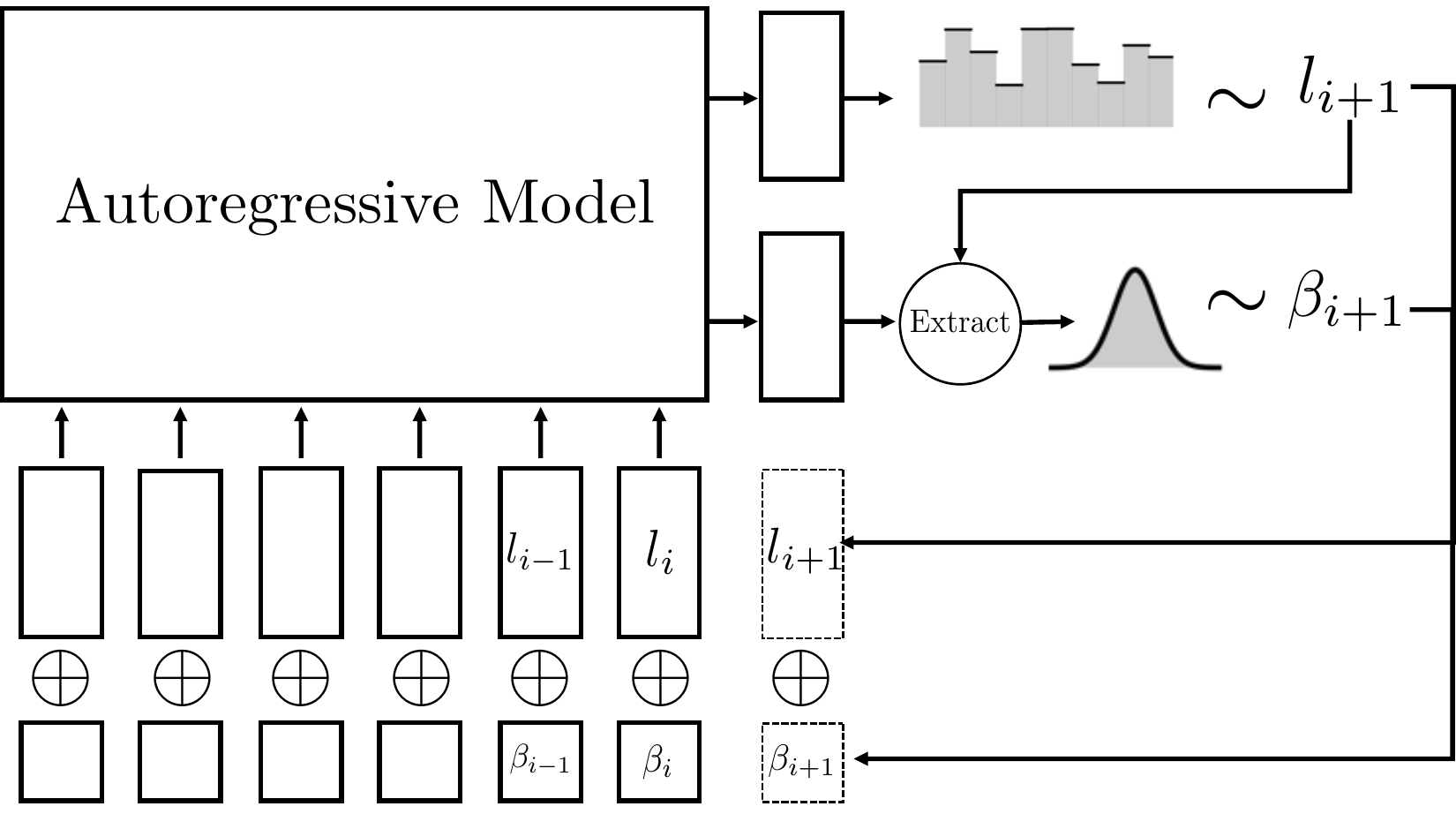}}
    \caption{Comparison of the standard decoupled approach and \ALGNAME for discrete-continuous optimization using an autoregressive model.
    In the decoupled approach, the discrete skeleton $\tau_{\text{d}} = \langle (l_1, \cdot), \ldots, (l_T, \cdot) \rangle$ is sampled first
    and then the continuous parameters $\beta_1, \ldots, \beta_T$ are optimized independently.
    In contrast, \ALGNAME models the joint distribution
    over the sequence of tokens $\langle (l_1, \beta_1), \ldots, (l_T, \beta_T)\rangle$.
    Here, the notation $\oplus$ stands for concatenation of vectors.}
    \end{center} 
\end{figure}

Deep learning methods have shown success in important combinatorial optimization problems \citep{bello2016neural}, 
including generating interpretable policies for continuous control
\citep{landajuela2021discovering} and symbolic regression (SR) to discover the underlying mathematical equations from the data
\citep{petersen2021deep,biggio2021neural,kamienny2022end, landajuela2022unified}.
Existing approaches train a generative model that constructs a solution to the optimization problem by sequentially choosing from a set of discrete tokens, using the value of the objective function as the terminal reward for learning.
However, these approaches do not jointly optimize the discrete and continuous components of such hybrid problems: Certain discrete tokens require the additional specification of an associated real-valued parameter, such as the threshold value at a decision tree node or the value of a constant token in an equation, but the learned generative model does not produce these values.
Instead, they adopt the design choice of \textit{decoupled optimization}, whereby only the construction of a discrete solution skeleton is optimized by deep learning, while the associated continuous parameters are left to a separate black-box optimizer.

We hypothesize that a joint discrete-continuous optimization approach (\Cref{fig:continuous-discrete-disco}) that generates a complete solution based on 
deep reinforcement learning (RL) \citep{sutton2018reinforcement} has significant advantages compared to existing decoupled approaches 
that employ learning only for the discrete skeleton (\Cref{fig:continuous-discrete-decoupled}).
In terms of efficiency, a joint approach only requires one evaluation of the objective function for each candidate solution,
whereas the decoupled approach based on common non-linear black-box optimization methods such as BFGS \citep{fletcher2000practical}, simulated annealing \citep{xiang1997generalized}, or evolutionary algorithms \citep{storn1997differential} requires a significant number of function evaluations to optimize each discrete skeleton.
This decoupled approach incurs a high cost for applications such as interpretable control, where each objective function evaluation involves running the candidate solution on many episodes of a high-dimensional and stochastic physical simulation \citep{landajuela2021discovering}.
Furthermore, joint exploration and learning on the full discrete-continuous solution space has the potential to escape from local optima and use information from prior samples to guide the subsequent search.

In this work, we consider discrete-continuous optimization problems that exhibit several key distinguishing features:
(1) a black-box reward,
(2) a variable-length structure of the design space, and
(3) a sequential structure in the form of \emph{prefix-dependent positional constraints}.
These problems are not well-suited to existing joint optimization approaches 
such as Mixed Integer Programming (MIP) \citep{fischetti2018deep, nair2021solving} or Mixed Bayesian Optimization (BO) \citep{daxberger2019mixed},
which are designed for problems with fixed-length discrete components
and do not naturally handle positional constraints in the design space.
To address these challenges, we draw upon the success of deep reinforcement learning in parameterized action 
space Markov decision processes \citep{hausknecht2015deep} to extend existing deep learning methods for 
discrete optimization \citep{bello2016neural,zophle2017nas,petersen2021deep,landajuela2021discovering} to the broader space of joint discrete-continuous optimization.
We summarize the main contributions of this paper as follows:
\begin{itemize}[leftmargin=*]
    \item We propose a novel method for joint discrete-continuous optimization using 
    autoregressive models and deep reinforcement learning, 
    which we call \ALGNAME, that is suited for black-box hybrid optimization problems
    over variable-length search spaces with prefix-dependent positional constraints.
    \item We present a novel formulation for decision tree policy search in control tasks as sequential discrete-continuous optimization and propose a method 
    for sequentially finding bounds for parameter ranges in decision nodes. 
    \item We perform exhaustive empirical evaluation of \ALGNAME on a diverse set of tasks, including interpretable control policies and symbolic regression. We show that \ALGNAME outperforms decoupled approaches on all tasks.
\end{itemize} 
\section{Related work}

\subsubsection{Hybrid discrete-continuous action spaces in reinforcement learning.}
The treatment of the continuous parameters as part of the action space
has strong parallels in the space of hybrid discrete-continuous RL.
In \citet{hausknecht2015deep}, the authors present a successful application of deep reinforcement learning to a domain with continuous state and action spaces.
In \citet{xiong2018parametrized}, the authors take an off-policy DQN-type approach that directly works on the hybrid
action space without approximation of the continuous part or relaxation of the discrete part, but
requires an extra loss function for the continuous actions.
In \citet{neunert2020continuous}, they propose a hybrid RL algorithm that uses continuous policies for discrete action selection and discrete policies for continuous action selection.

\subsubsection{Symbolic regression with constants optimization.}
In the field of symbolic regression, different approaches have been proposed for addressing the optimization of both discrete skeletons and continuous parameters.
Traditional genetic programming approaches and deep generative models handle these problems separately, with continuous constants optimized after discrete parameters \citep{topchy2001faster,petersen2021deep,biggio2021neural}. 
 Recent works aim to jointly optimize discrete constants and continuous parameters by relaxing the discrete problem into a continuous one \citep{martius2016extrapolation, sahoo2018learning}, or by tokenizing (i.e., discretizing) the continuous constants \citep{kamienny2022end}.
The former approach faces challenges such as exploding gradients and the need to revert continuous values to discrete ones.
The latter approach tokenizes continuous constants, treating them similarly to discrete tokens, but such quantization is problem-dependent, restricts the search space, and requires additional post-hoc optimization to refine the continuous parameters. 

\subsubsection{Decision tree policies in reinforcement learning.}
In the domain of symbolic reinforcement learning,
where the goal is to find intelligible and concise control policies,
works such as \citet{landajuela2021discovering} and \citet{sahoo2018learning}
have discretized the continuous space and used relaxation approaches, respectively,
to optimize  symbolic control policies in continuous action spaces.
For discrete action spaces, 
a natural representation of a symbolic policy is a \DT \citep{ding2020cdt,silva2020optimization,custode2023evolutionary}.
In \citet{custode2023evolutionary}, the authors use an evolutionary search to find the best \DT policy
and further optimized the real valued thresholds using a decoupled approach.
Relaxation approaches find their counterparts within this domain
in works such as \citet{sahoo2018learning,silva2020optimization, ding2020cdt},
where a soft \DT is used to represent the policy.
The soft \DT, which fixes the discrete structure of the policy and exposes the continuous parameters,
is then optimized using gradient-based methods.

\section{Discrete-Continuous Deep Symbolic Optimization}
\label{sec:neural-guided-search}

\subsection{Notation and problem definition}
\label{sec:notation}
We consider a discrete-continuous optimization problem defined over a search space $\mathcal{T}$ 
of sequences of tokens $\tau = \langle \tau_1, \ldots, \tau_T \rangle$,
where each token $\tau_i$ belongs to a library $\mathcal{L}$
and the length $T$ of the sequence is not fixed \textit{a priori}.
The library $\mathcal{L}$ is a set of $K$ tokens $\mathcal{L} = \{l_1, \ldots, l_{K}\}$,
where a subset $\hat{\mathcal{L}} \subseteq \mathcal{L}$ of them are parametrized by a continuous parameter, i.e.,
each token $l \in \hat{\mathcal{L}}$ has an associated continuous parameter $\beta \in \mathcal{A}(l) \subset \mathbb{R}$,
where $\mathcal{A}(l)$ is the token-dependent range.
To ease the notation, 
we define $\bar{\mathcal{L}} \defeq \mathcal{L} \setminus \hat{\mathcal{L}}$ and
consider a dummy range $\mathcal{A}(l) = [0,1]\subset \mathbb{R}$ for the
strictly discrete tokens $l \in \bar{\mathcal{L}}$. Thus, we define
$$
l(\beta) = \begin{cases}
l & \text{if }  l \in \bar{\mathcal{L}} \\
l(\beta) & \text{if } l \in \hat{\mathcal{L}}
\end{cases}, \forall (l, \beta) \in \mathcal{L} \times \mathcal{A}(l).
$$
In other words, the parameter $\beta$ is ignored if $l \in \bar{\mathcal{L}}$.
With this notation, 
we can write $\tau_i = l_i(\beta_i) \in \mathcal{L}, \forall i \in \{1, \ldots, T\}$.
In the following, we use the notation $l_i(\beta_i) \equiv (l_i, \beta_i)$
and write
$
    \tau = \langle\tau_1, \ldots, \tau_T\rangle = \langle l_1(\beta_1), \ldots, l_T(\beta_T)\rangle 
    \equiv \langle (l_1, \beta_1), \ldots, (l_T, \beta_T)\rangle.
$

Given a sequence $\tau$, we define the \emph{discrete skeleton} $\tau_{\text{d}}$ as the sequence obtained by removing the continuous parameters from $\tau$, i.e., $\tau_{\text{d}} = \langle (l_1, \cdot), \ldots, (l_T, \cdot)\rangle.$
We introduce the operator ${\tt eval} : \mathcal{T}  \rightarrow \mathbb{T}$ to represent the
semantic interpretation of the sequence $\tau$ as an object in the relevant design space $\mathbb{T}$. We consider problems with \emph{prefix-dependent positional constraints}, i.e.,
problems for which, given a prefix $\tau_{1:(i-1)}$, 
there exists a possible non-empty set of unfeasible tokens $\mathcal{C}_{\tau_{1:(i-1)}} \subseteq \mathcal{L}$
such that ${\tt eval}(\tau_{1:(i-1)} \cup \tau_i \cup \tau_{(i+1):T}) \notin \mathbb{T}$ for all $\tau_i \in \mathcal{C}_{\tau_{1:(i-1)}}$
and for all $\tau_j \in \mathcal{L}$ with $i < j \leq T$.
Variable-length problems exhibiting such constraints are not 
well-suited for MIP solvers or classical Bayesian Optimization methods.

The optimization problem is defined by the reward function $R : \mathbb{T} \rightarrow \mathbb{R}$,
which can be deterministic or stochastic.
In the stochastic case, we have a reward distribution 
$p_R(r | t)$ conditioned 
on the design $t \in \mathbb{T}$
and the reward function is given by
$
R(t) = \mathbb{E}_{r \sim p_R(r | t)}[r].
$
Note that we do not assume that the reward function $R$ is differentiable with respect to the continuous parameters $\beta_i$.
In the following, we make a slight abuse of notation and use 
$R(\tau)$ and $p_R(r | \tau)$ 
to denote $R({\tt eval}(\tau))$ and $p_R(r | {\tt eval}(\tau))$, respectively.
The optimization problem
is to find a sequence
$\tau^* = \langle\tau^*_1, \ldots, \tau^*_T\rangle
= \langle (l^*_1, \beta^*_1), \ldots, (l^*_T, \beta^*_T)\rangle
$
(where the length $T$ is not fixed a priori)
such that 
$\tau^* \in \argmax_{\tau \in \mathcal{T}} R(\tau)$.

\subsection{Method}
\label{sec:method}

\subsubsection{Combinatorial optimization with autoregressive models.}
In applications of deep learning to combinatorial optimization \citep{bello2016neural},
a probabilistic model $p(\tau)$ is learned over the design space $\mathcal{T}$.
The model is trained to gradually allocate more probability mass to high scoring solutions.
The training can be done using supervised learning, 
if problem instances with their corresponding solutions are available,
or,
more generally,
using RL.
In most cases, the model $p(\tau)$ is parameterized by 
an autoregressive (\NN) model with parameters $\theta$.
The model is used to generate sequences as follows.

At position $i$,
the model emits a vector of logits $\psi^{(i)}$
conditioned on the previously generated tokens $\tau_{1:(i-1)}$, i.e., $\psi^{(i)} = \NN(\tau_{1:(i-1)}; \theta)$.
The new token $\tau_i$ is sampled from the distribution
$
p(\tau_i | \tau_{1:(i-1)}, \theta) = \text{softmax}(\psi^{(i)})_{\mathcal{L}(\tau_i)},
$
where $\mathcal{L}(\tau_i)$ is the index in $\mathcal{L}$ corresponding to node value $\tau_i$.
The new token $\tau_i$ is then added to the sequence $\tau_{1:(i-1)}$ and used to condition the generation of the next token $\tau_{i+1}$.
The process continues until a stopping criterion is met.

Different model architectures can be employed to generate the logits $\psi^{(i)}$. For instance, 
recurrent neural networks (RNNs) have been utilized in \citet{petersen2021deep,landajuela2021discovering,mundhenk2021symbolic, da2023language}, 
and transformers with causal attention have been applied in works like \citet{biggio2021neural} and \citet{kamienny2022end}.

\subsubsection{Prefix-dependent positional constraints.}
Sequential token generation enables flexible configurations and the incorporation of constraints during the search process \citep{petersen2021deep}. 
Specifically, 
given a prefix $\tau_{1:(i-1)}$,
a prior {\scriptsize $\psi^{(i)}_{\circ} \in \mathbb{R}^{|\mathcal{L}|}$} is computed
such that {\scriptsize ${\psi^{(i)}_{\circ}}_{\mathcal{L}(\tau_i)} = -\infty$} for tokens $\tau_i$ in the unfeasible set $\mathcal{C}_{\tau_{1:(i-1)}}$
and zero otherwise.
The prior is added to the logits $\psi^{(i)}$ before sampling the token $\tau_i$.

\subsubsection{Extension to discrete-continuous optimization.}
Current deep learning approaches for combinatorial optimization
only support discrete tokens, i.e., $\hat{\mathcal{L}} = \emptyset$,
\citep{bello2016neural}
or completely decouple the discrete and continuous parts of the problem, as in
\citet{petersen2021deep, landajuela2021discovering, mundhenk2021symbolic, da2023language},
by sampling first the discrete skeleton $\tau_{\text{d}}$ and then
optimizing its continuous parameters separately (see \Cref{fig:continuous-discrete-decoupled}).
In this work, we extend these frameworks to support
\emph{joint optimization of discrete and continuous tokens}.
The model is extended to emit two outputs $\psi^{(i)}$ and $\phi^{(i)}$ for each token $\tau_i = (l_i, \beta_i)$
conditioned on the previously generated tokens, i.e.,
$\left( \psi^{(i)}, \phi^{(i)} \right) = \NN((l, \beta)_{1:(i-1)}; \theta)$,
where we use the notation $(l, \beta)_{1:(i-1)}$ to denote the sequence of tokens $\langle(l_1, \beta_1), \ldots, (l_{i-1}, \beta_{i-1})\rangle$
(see \Cref{fig:continuous-discrete-disco}).
Given 
tokens $(l, \beta)_{1:(i-1)}$, the $i^\textrm{\tiny th}$ token $(l_i, \beta_i)$
is generated by sampling from the following distribution:
$$
 p((l_i, \beta_i) | (l, \beta)_{1:(i-1)}, \theta) = 
\begin{cases}
    \mathcal{U}_{[0,1]}(\beta_i) \text{softmax}(\psi^{(i)})_{\mathcal{L}(l_i)} & \text{if } l_i \in \bar{\mathcal{L}} \\
    \mathcal{D}(\beta_i | l_i, \phi^{(i)}) \text{softmax}(\psi^{(i)})_{\mathcal{L}(l_i)} & \text{if } l_i \in \hat{\mathcal{L}} \\
\end{cases},
$$
where $\mathcal{D}(\beta_i | l_i, \phi^{(i)})$ is the probability density function of the distribution $\mathcal{D}$ that is used to sample $\beta_i$ from $\phi^{(i)}$.
Note that the choice of $\beta_i$ is conditioned on the choice of discrete token $l_i$.
We assume that the support of $\mathcal{D}(\beta|l,\phi)$ is a subset of $\mathcal{A}(l)$ for all $l \in \hat{\mathcal{L}}$. 
Additional priors of the form 
$(\psi^{(i)}_{\circ}, 0)$
can be added to the logits before sampling the token $\tau_i$.

\def\distok{l}
\def\contok{\beta}
\def\paramtokset{\hat{\mathcal{L}}}
\def\distokset{\bar{\mathcal{L}}}

\subsubsection{Training \ALGNAME.}
The parameters $\theta$ of the model are learned by maximizing the expected reward
$J(\theta) = \mathbb{E}_{\tau \sim p(\tau | \theta)} [R(\tau)]$
or, alternatively, the quantile-conditioned expected reward
$
J_{\varepsilon}(\theta) = \mathbb{E}_{\tau \sim p(\tau | \theta)} [R(\tau) | R(\tau) \geq R_{\varepsilon}(\theta)],
$
where $R_{\varepsilon}(\theta)$ represents the $(1- \varepsilon)$-quantile of the reward distribution $R(\tau)$ 
sampled from the trajectory distribution $p(\tau | \theta)$.
The motivation for using $J_{\varepsilon}(\theta)$ is to encourage the model 
to focus on \emph{best case} performance over \emph{average case} performance (see \citet{petersen2021deep}),
which is the preferred behavior in optimization problems.
It is worth noting that both objectives, 
$J(\theta)$ and $J_{\varepsilon}(\theta)$, serve as \emph{relaxations} of the original $\argmax R(\tau)$ optimization problem described above.

To optimize the objective $J_{\varepsilon}(\theta)$, 
we extend the risk-seeking policy gradient of \citet{petersen2021deep}
to the discrete-continuous setting.
The gradient of $J_{\varepsilon}(\theta)$ reads as
\begin{align*}
    \nabla_{\theta} J_{\varepsilon}(\theta) = \mathbb{E}_{\tau \sim p(\tau | \theta)} \left[ 
    A(\tau, \varepsilon, \theta) \, S((l, \beta)_{1:T})
    \mid  A(\tau, \varepsilon, \theta) >0 \right],
\end{align*}
where $A(\tau, \varepsilon, \theta) = R(\tau) - R_{\varepsilon}(\theta)$ and 
\begin{align*}
    S((l, \beta)_{1:T}) = \displaystyle\sum_{i=1}^{T}
    \begin{cases}
    \nabla _{\theta }\log p(l_i | (l, \beta)_{1:(i-1)}, \theta) & \text{if } l_i \in \bar{\mathcal{L}}, \\[2ex]
    \begin{aligned}
    &\nabla_\theta \log p(l_i | (l, \beta)_{1:(i-1)}, \theta) \\
    &\ \ \ \ \ \ \ \ + \nabla_\theta \log p(\beta_i | l_{1:i}, \beta_{1:i-1}, \theta)
    \end{aligned} & \text{if } l_i \in \hat{\mathcal{L}}.
    \end{cases}
\end{align*}
We provide pseudocode for \ALGNAME,
a derivation of the risk-seeking policy gradient, and additional details of the learning procedure in the appendix.
\section{Experiments}
\label{sec:experiments}

We demonstrate the benefits and generality of our approach on a diverse set of tasks as follows.
Firstly, we introduce a new pedagogical task, called \textit{Parameterized Bitstring}, to understand the conditions under which the benefits of \ALGNAME versus decoupled approaches become apparent.
We then consider two preeminent tasks in combinatorial optimization:
\DT policy optimization for reinforcement learning and symbolic regression for equation discovery.

\subsubsection{Baselines.} To demonstrate the advantages of joint discrete-continuous optimization, we compare \ALGNAME with the following classes of methods:
\begin{itemize}[leftmargin=*]
    \item \textbf{Decoupled-RL-}\{\textbf{BFGS}, \textbf{anneal}, \textbf{evo}\}:
    This baseline trains a generative model with reinforcement learning to produce a discrete skeleton \citep{petersen2021deep}, 
    which is then optimized by a downstream nonlinear solver for the continuous parameters.
    The objective value at the optimized solution is the reward, which is used to update the generative model using the same policy gradient approach and architecture as DisCo-DSO.
    The continuous optimizer is either L-BFGS-B (BFGS), simulated annealing (anneal) \citep{xiang1997generalized}, or differential evolution (evo) \citep{storn1997differential}, using the SciPy implementation \citep{2020SciPy-NMeth}.
    \item \textbf{Decoupled-GP-}\{\textbf{BFGS}, \textbf{anneal}, \textbf{evo}\}:
    This baseline uses genetic programming (GP) \citep{koza1990genetic} to produce a discrete skeleton,
    which is then optimized by a downstream nonlinear solver for the continuous parameters.
    \item \textbf{BO}:
    For the Parameterized Bitstring task, 
    which has a fixed length search space and no positional constraints, 
    we also consider a Bayesian Optimization baseline using
    expected improvement as acquisition function \citep{shahriari2015taking, garrido2020dealing}.
\end{itemize}

All experiments involving RL and \ALGNAME use a RNN 
with a single hidden layer of 32 units as the generative model.
The GP baselines use the ``Distributed Evolutionary Algorithms in Python''
software\footnote{\url{https://github.com/DEAP/deap}. LGPL-3.0 license.}
\citep{fortin2012deap}. 
Additional details are provided in the appendix.

\subsubsection{Note on baselines for symbolic regression.}
In the context of symbolic regression, 
some of the above baselines corresponds to popular methods in the literature. Specifically,
\textbf{Decoupled-RL-BFGS} corresponds exactly to the method ``Deep Symbolic Regression'' from \citet{petersen2021deep},
and \textbf{Decoupled-GP-BFGS} corresponds to a standard implementation of genetic programming for symbolic regression \textit{à la} \citet{koza1994genetic}
(most common approach to symbolic regression in the literature).

\subsection{Parameterized bitstring task}
\label{sec:parameterized_bitstring}

\subsubsection{Problem formulation.}
We design a general and flexible \textit{Parameterized Bitstring} benchmark problem, denoted $\text{PB}(N, f, l^*, \beta^*)$, to test the hypothesis that \ALGNAME is more efficient than the decoupled optimization approach.
In each problem instance, the task is to recover a hidden string $l^* \in [0,1]^T$ of $T$ bits and a vector of parameters $\beta^* \in \Rbb^T$.
Each bit $l^*_i$ is paired with a parameter $\beta^*_i$ via the reward function $R$, which gives a positive value based on an objective function $f(\beta_i, \beta^*_i) \in [0,1]$ only if the correct bit $l_i^*$ is chosen at position $i$:
\begin{align} \label{eq:bs_reward}
    R(\tau, \beta) \defeq \frac{1}{T} \sum_{i=1}^T \boldsymbol{1}_{\tau_i = \tau^*_i} \left( \alpha + (1-\alpha) f(\beta_i, \beta^*_i)\right)
\end{align}


The scalar $\alpha \in [0,1]$ controls the relative importance of expending computational effort to optimize the discrete or continuous parts of the reward.
The problem difficulty can be controlled by increasing the length $T$ and increasing the nonlinearity of the objective function $f$, such as by increasing the number of local optima.
In our experiment, we tested the following objective functions, which represent objectives with multiple suboptimal local maxima ($f_1$) and discontinuous objective landscapes ($f_2$):
\begin{align}
    f_1(x, x^*) \defeq \left\lvert \frac{\sin(50(x - x^*))}{50(x - x^*)} \right\rvert, \quad \quad \\
    f_2(x, x^*) \defeq 
    \begin{cases}
        1, \quad &|x - x^*| \leq 0.05 \\
        0.5, \quad &0.05 < |x - x^*| \leq 0.1 \\
        0, \quad &0.1 < |x - x^*|
    \end{cases}.
    \label{eq:bitstring_error_functions}
\end{align}

\subsubsection{Results.}
\begin{figure}[h!]
    \centering
    \subfigure[Parameterized Bitstring with $f_1$]{\includegraphics[width=0.48\linewidth]{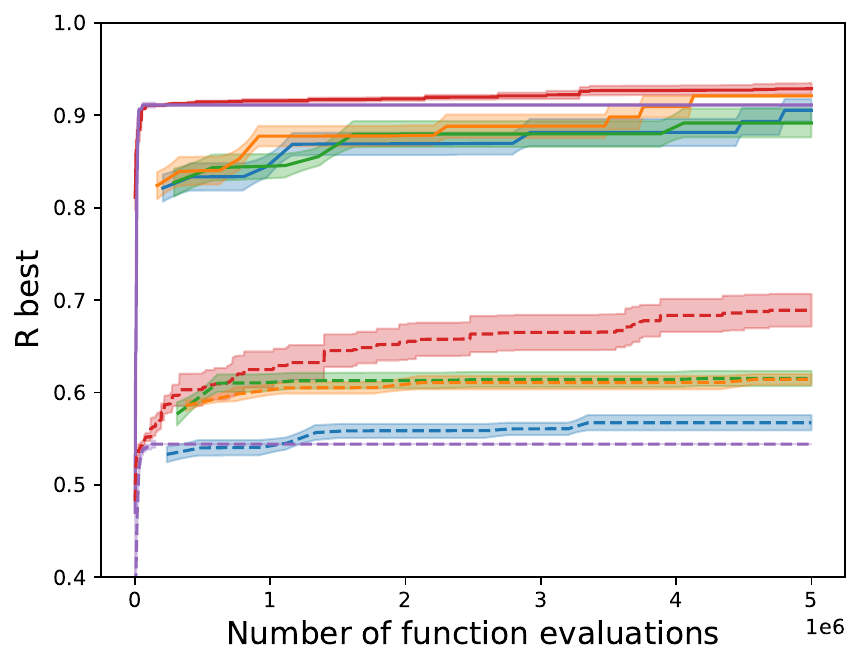}} \hfil
    \subfigure[Parameterized Bitstring with $f_2$]{\includegraphics[width=0.48\linewidth]{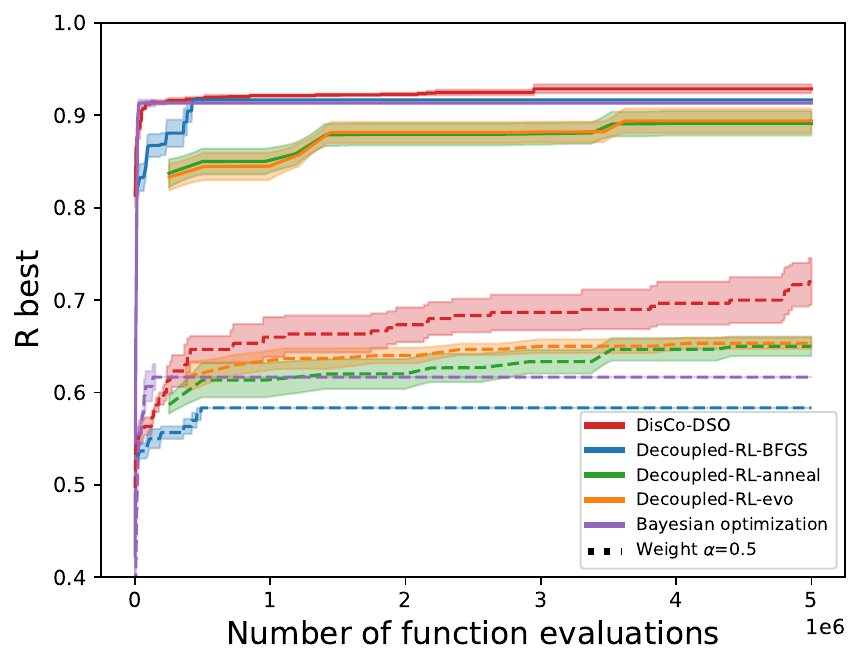}}
    \caption{Reward of best solution versus number of function evaluations on a parameterized bitstring task, for two continuous optimization landscapes $f_1$ and $f_2$ and weights $\alpha = 0.5, 0.9$. Solid line corresponds to weight $\alpha=0.9$, dashed line $\alpha=0.5$. Mean and standard error over 5 seeds.}
    \label{fig:param_bitstring_rbest}
\end{figure}

\Cref{fig:param_bitstring_rbest} shows that \ALGNAME is significantly more sample efficient than the decoupled approach when the discrete solution contributes more to the overall reward.
This is because each sample generated by \ALGNAME is a complete solution, which costs only one function evaluation to get a reward.
In contrast, each sample generated by the baseline decoupled methods only has a discrete skeleton, which requires many function evaluations using the downstream optimizer to get a single complete solution.
As the discrete skeleton increases in importance, the relative contribution of function evaluations for continuous optimization decreases.
Note that, given the same computational budget, the
BO method performs less function evaluations than the rest of the methods
and the final results are worse than \ALGNAME.
This is because BO has a computational complexity of $\mathcal{O}(n^3)$ \citep{shahriari2015taking}, 
where $n$ is the number of function evaluations.
This computational complexity
makes BO challenging or even infeasible
for large $n$ \citep{lan2022time}.

\subsection{Decision tree policies for reinforcement learning} 
\label{sec:control}

\subsubsection{Problem formulation.}
In this section we consider the problem of discovering \DT policies for RL.
We consider $\mathbb{T}$ as the space of univariate \DTs \citep{silva2020optimization}.
Extensions to multivariate \DTs, also known as oblique trees,
are possible, but we leave them for future work.
Given an RL environment with observations $x_1, \dots, x_n$ and discrete actions $a_1, \dots, a_m$, 
we consider the library of Boolean expressions and actions given by
$
\mathcal{L} = \{x_1 < \beta_{1}, \dots, x_n < \beta_{n}, a_1, \dots, a_m\},
$
where $\beta_1, \dots, \beta_n$ are the values of the observations that are used in the internal nodes of the \DT.
The evaluation operator ${\tt eval} : \mathcal{T} \rightarrow \mathbb{T}$ is defined as follows.
We treat sequence $\tau$ as the pre-order traversal of a decision tree,
where the decision tokens ($x_n < \beta_n$) are treated as binary nodes 
and the action tokens ($a_n$) are treated as leaf nodes.
For evaluating the decision tree,
we start from the root node 
and follow direction
\begin{equation*}
  D_{x_n < \beta_n} (x) = \begin{cases}
      \text{left if } x_n < \beta_n \text{ is True,} \\
      \text{right if } x_n < \beta_n \text{ is False,}
  \end{cases}
\end{equation*}
for every decision node encountered until we reach a leaf node.
See \Cref{fig:decision-tree-parameter-bounds} for an example.
The reward function is defined as 
$R(t) = \mathbb{E}_{r \sim p_R(r | t)}[r]$
where $p_R(r | t)$ is the reward distribution following policy $t$ in the environment.
In practice, we use the average reward over $N$ episodes, i.e., 
$
R(t) = \frac{1}{N} \sum_{i=1}^N r_i,
$
where $r_i$ is the reward obtained in episode $i$.
Prefix-dependent positional constraints for this problem 
are given in the appendix.

\subsubsection{Sampling decision nodes in decision trees.}
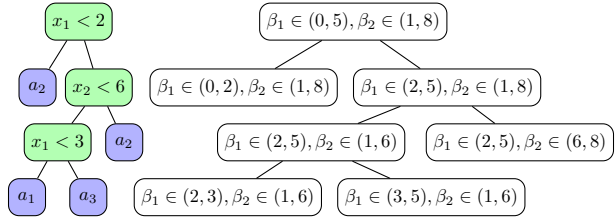
\begin{figure}[h]
\begin{center}
\begin{tikzpicture}[
  every node/.style = {shape=rectangle, rounded corners, draw, align=center, minimum size=2em, inner sep=1pt, scale=0.7}, 
  level 1/.style = {level distance = 12mm}, 
  level 2/.style = {level distance = 10mm}, 
  scale=0.73 
]

\node at (0, 0) [fill=green!30]{{$\;x_1 < 2\;$}}
  child{ node[black, fill=blue!30, xshift=0mm, solid]{{$a_2$}} }
  child{ node[black, fill=green!30, xshift=-4mm, solid]{$\;x_2 < 6\;$} edge from parent [solid, black]
    child{ node[black, fill=green!30, xshift=0mm, solid]{$\;x_1 < 3\;$}
      child { node[fill=blue!30,xshift=2mm]{{$a_1$}} }
      child { node[fill=blue!30,xshift=-2mm]{{$a_3$}} }
    }
    child { node[fill=blue!30,xshift=-3mm]{{$a_2$}} }
  };

\node at (5, 0) []{{$\; \beta_1 \in (0, 5), \beta_2 \in (1, 8) \;$}} 
  child{ node[xshift=-13mm]{{$\; \beta_1 \in (0, 2), \beta_2 \in (1, 8) \;$}} }
  child{ node[xshift=10mm]{$\; \beta_1 \in (2, 5), \beta_2 \in (1, 8) \;$} edge from parent [solid, black]
    child{ node[xshift=-18mm]{$\; \beta_1 \in (2, 5), \beta_2 \in (1, 6) \;$}
      child{ node[xshift=-8mm]{{$\; \beta_1 \in (2, 3), \beta_2 \in (1, 6) \;$}} }
      child{ node[xshift=15mm]{{$\; \beta_1 \in (3, 5), \beta_2 \in (1, 6) \;$}} }
    }
    child{ node[xshift=6mm]{{$\; \beta_1 \in (2, 5), \beta_2 \in (6, 8) \;$}} }
  };

\end{tikzpicture}
\end{center}
\caption{Left: the 
\DT
associated with the traversal $\langle x_1<2, a_2, x_2 < 6, x_1 < 3, a_1, a_3, a_2 \rangle$. 
Right: the corresponding bounds for the parameters during the sampling process (suppose the bounds for observations $x_1$ and $x_2$ are respectively [0, 5] and [1, 8]).}
\label{fig:decision-tree-parameter-bounds}
\end{figure}

To efficiently sample decision nodes, we employ truncated normal distributions to select parameters $\beta_i$ within permissible ranges. Many RL environments place boundaries on observations, and the use of the truncated normal distribution guarantees that parameters will only be sampled within those boundaries. Additionally, a decision node which is a child of another decision node cannot select parameters from the environment-enforced boundaries. This is because the threshold involved at a decision node changes the range of values which will be observed at subsequent decision nodes. In this way, a previous decision node "dictates" the bounds on a current decision node.
For instance, consider the \DT displayed in Figure~\ref{fig:decision-tree-parameter-bounds}. Assume that 
the observation $x_1$ falls within the interval $[0, 5]$ (note that in practice the RL environment provided bounds are used to determine the interval), and the tree commences with the node $x_1 < 2$. In the left child node, as $x_1 < 2$ is true, there is no need to evaluate whether $x_1$ is less than 4 (or any number between 2 and 5), as that is already guaranteed. Consequently, we should sample a parameter $\beta_1$ within the range (0, 2). Simultaneously, since we do not assess the Boolean expression regarding $x_2$, the bounds on $\beta_2$ remain consistent with those at the parent node. The parameter bounds for the remaining nodes are illustrated in Figure~\ref{fig:decision-tree-parameter-bounds}. The procedure for determining these maximum and minimum values is outlined in Algorithm~\ref{alg:decision-tree-parameter-bounds} in Appendix~\ref{app:priors}.

\begin{figure}[h]
    \centering
    \subfigure[Acrobot-v1]{\includegraphics[width=0.48\linewidth]{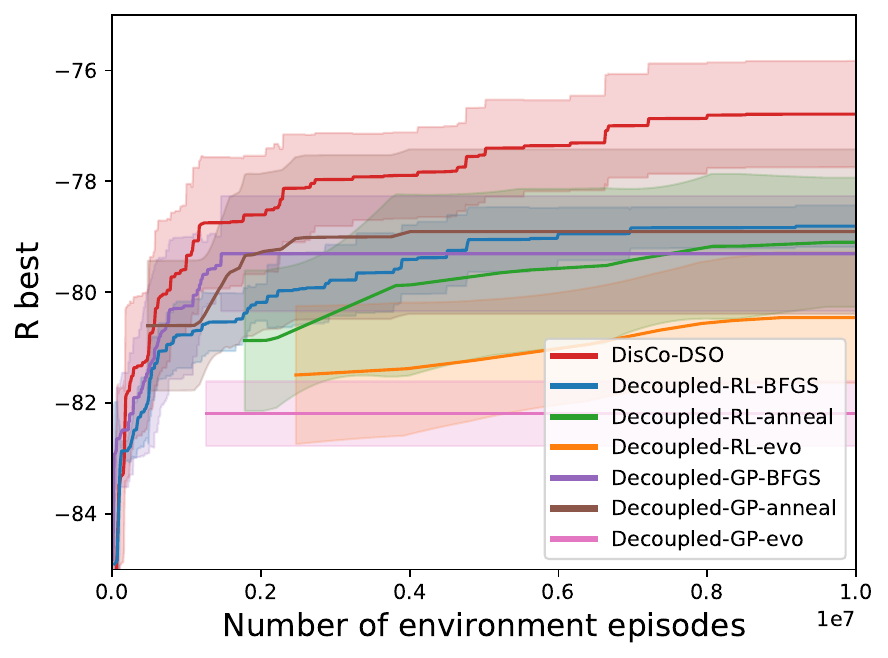}} \hfil
    \subfigure[LunarLander-v2]{\includegraphics[width=0.48\linewidth]{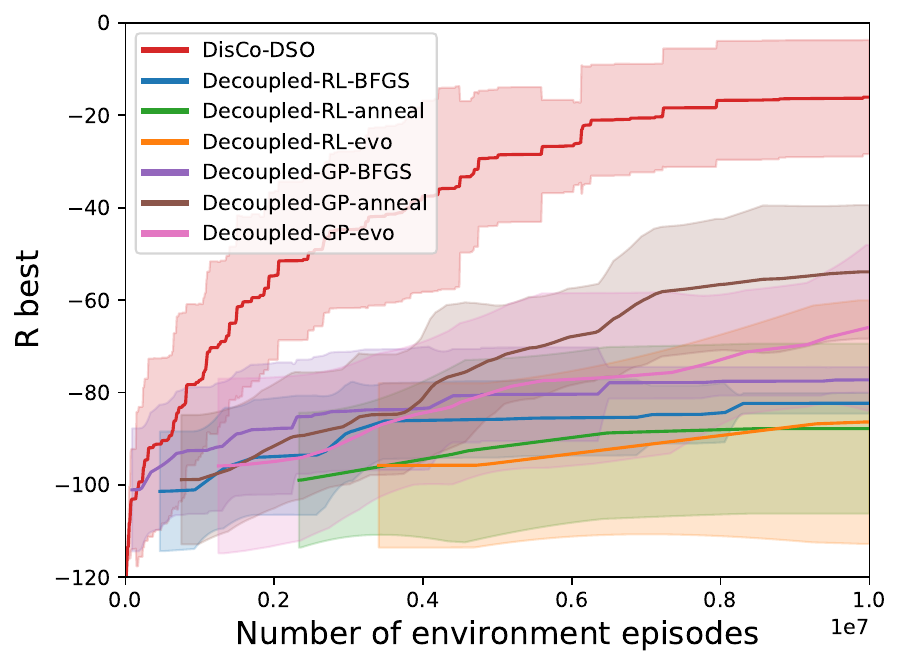}}
    \caption{Reward of the best solution versus number of function evaluations on the \DT policy task, for Acrobot-v1 and LunarLander-v2.}
    \label{fig:learning_curves}
\end{figure}

\subsubsection{Evaluation.}
For evaluation, we follow other works in the field \citep{silva2020optimization, ding2020cdt, custode2023evolutionary}
and use the OpenAI Gym's \citep{1606.01540} environments MountainCar-v0, CartPole-v1, Acrobot-v1, and LunarLander-v2.
We investigate the sample-efficiency of \ALGNAME on the \DT policy task
when compared to the decoupled baselines described at the beginning of this section.
We train each algorithm for 10 different random seeds.

\subsubsection{Results.}
In \Cref{fig:learning_curves} (see also \Cref{fig:learning_curves_mc_cp} in the appendix),
we report the 
mean and standard deviation of the
best reward found by each algorithm
versus number of environment episodes.
These results show that \ALGNAME dominates the baselines in terms of sample-efficiency.
The trend is consistent across all environments, and is more pronounced in the more complex environments.
The efficient use of evaluations by \ALGNAME (each sample is a complete well-defined \DT)
versus the decoupled approaches, where each sample is a discrete skeleton that requires many evaluations to get a single complete solution,
becomes a significant advantage in the RL environments where each evaluation involves running the environment for $N$ episodes.

\subsubsection{Literature comparisons.}
We conduct a performance comparison of \ALGNAME against various baselines in the literature, namely evolutionary \DTs as detailed in \cite{custode2023evolutionary}, cascading \DTs introduced in \cite{ding2020cdt}, and interpretable differentiable \DTs (DDTs) introduced in \cite{silva2020optimization}.
In addition, we provide results with a BO baseline, where the structure of the \DT is fixed to a binary tree of depth 4 without prefix-dependent positional constraints.
Whenever a method provides a tree structure for a specific environment, we utilize the provided structure and assess it locally. In cases where the method's implementation is missing, we address this by leveraging open-source code. This approach allows us to train a tree in absent environments, ensuring that we obtain a comprehensive set of results for all methods evaluated across all environments. 
The \DTs found by \ALGNAME are shown in \Cref{fig:best_decision_trees} 
(see also \Cref{fig:best_decision_trees_mc_cp} in the appendix).
Comparisons are shown in \Cref{tab:control_results}.
Methods we trained locally are marked with an asterisk (*). Critically, we ensure consistent evaluation across baselines by assessing each decision tree policy on an identical set of 1,000 random seeds per environment.

\begin{figure}
\centering
\subfigure[Acrobot-v1]{\includegraphics[width=0.48\textwidth]{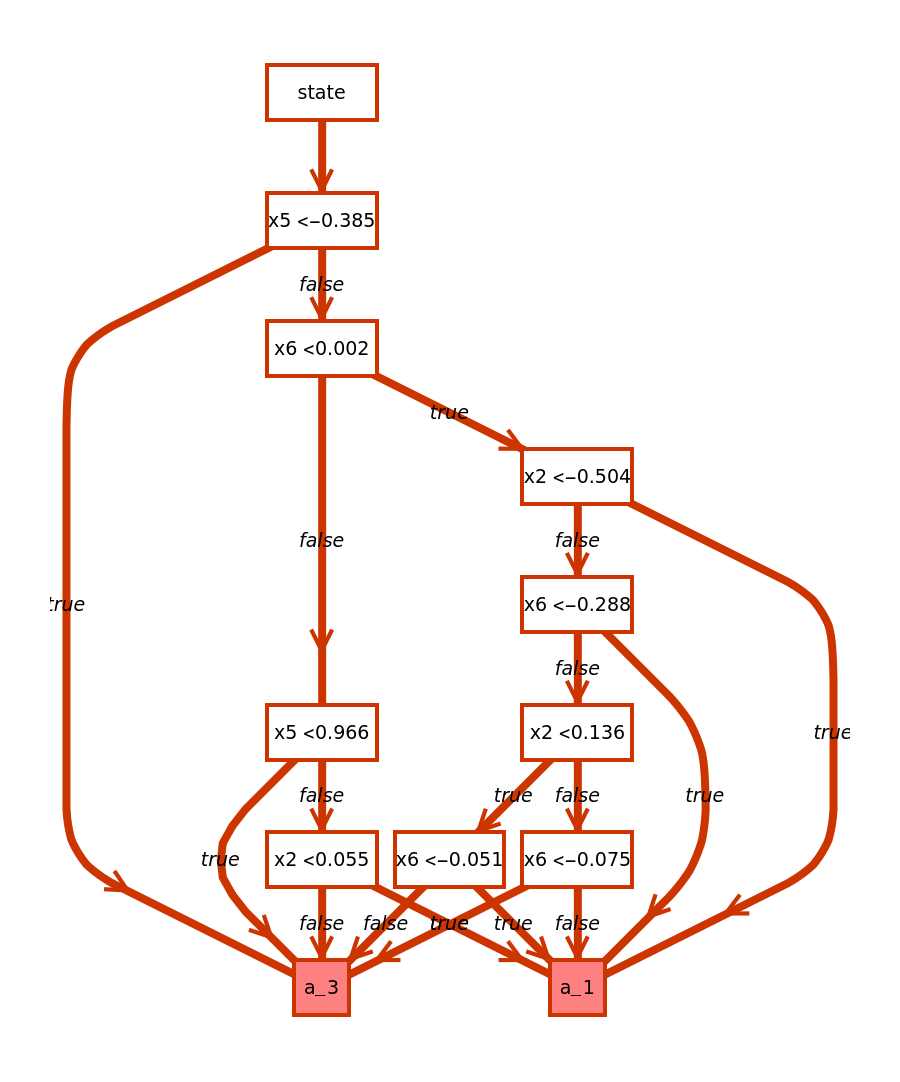}} \hfil
\subfigure[LunarLander-v2]{\includegraphics[width=0.4\textwidth]{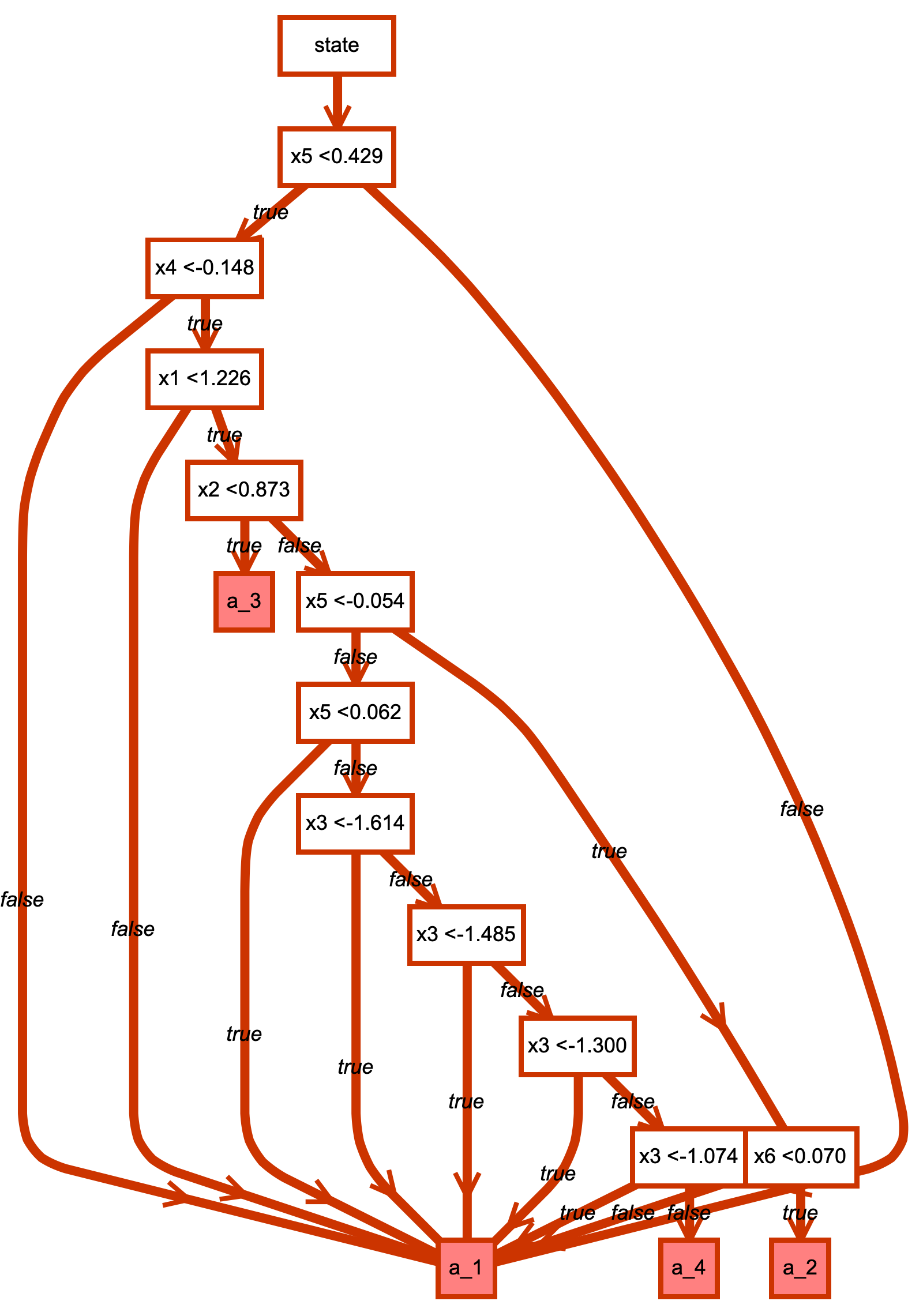}}
\caption{Best \DTs found by \ALGNAME on the \DT policy tasks for Acrobot-v1 and LunarLander-v2.}
\label{fig:best_decision_trees}
\end{figure}

\begin{table}[h!]
  \begin{center}
  \scalebox{0.8}{
  \centering
  \begin{tabular}{l|c|c|c|c|c|c|c|c}
      \multirow{2}{*}{Algorithm} &
      \multicolumn{2}{c|}{Acrobot-v1} &
      \multicolumn{2}{c|}{CartPole-v1}  &
      \multicolumn{2}{c|}{LunarLander-v2} & 
      \multicolumn{2}{c}{MountainCar-v0} \\
      \cline{2-9}
       &  MR & PC & MR & PC & MR & PC & MR & PC\\
        \hline
        DisCo-DSO & \textbf{-76.58} & 18 & \textbf{500.00} & 14 & \textbf{99.24} & 23 & \textbf{-100.97} & 15  \\
        Evolutionary DTs & -97.12* & 5 & 499.58 & 5 & -87.62* & 17 & -104.93 & 13 \\
        Cascading DTs  & -82.14* & 58 &496.63 & 22 & -227.02 & 29 & -200.00 & 10 \\
        Interpretable DDTs & -497.86* & 15 &389.79 & 11 & -120.38 & 19 & -172.21* & 15 \\
        Bayesian Optimization$^{\dagger}$ & -90.99* & 7 & 85.47* & 7 & -112.14* & 7 & -200.0* & 7 \\
  \end{tabular}
  }
  \end{center}
  \caption{Evaluation of the best univariate \DTs found by \ALGNAME and other baselines on the \DT policy task. Here, MR is the mean reward earned in evaluation over a set of 1,000 random seeds, while PC represents the parameter count in each tree. 
For models trained in-house (*), the figures indicate the parameter count after the discretization process.
$^{\dagger}$The topology of the tree is fixed for BO.
  }
  \label{tab:control_results}
\end{table}

In \Cref{tab:control_results} we also show the complexity of the discovered \DT
as measured by the number of parameters in the tree. We count every (internal or leaf) node of univariate \DTs (produced by all methods except for Cascading \DTs) as one parameter. For Cascading \DTs, the trees contain feature learning trees and decision making trees. The latter is just univariate \DTs, so the same complexity measurement is used. For the leaf nodes of feature learning trees, the number of parameters is number of observations times number of intermediate features.
From \Cref{tab:control_results}, we observe that the univariate \DTs
found by \ALGNAME have the best performance on all environments
at a comparable or lower complexity than the other literature baselines.


\subsection{Symbolic regression for equation discovery}
\label{sec:symbolic_regression}

\begin{figure}[t]
    \centering
    \includegraphics[width=1.0\linewidth]{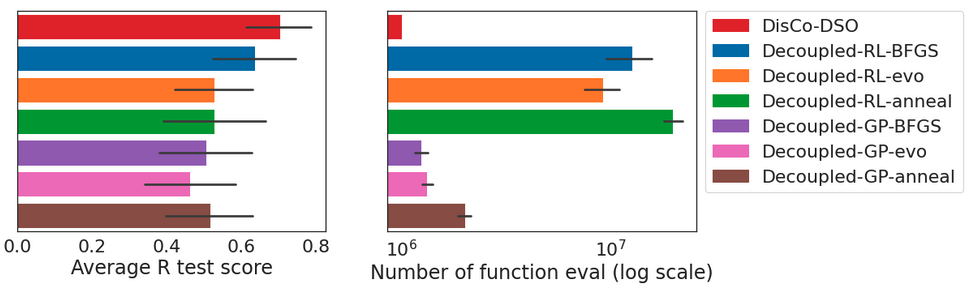}
    \caption{Average test set reward (left) and number of function evaluations (right) used across methods on the symbolic regression task.
    Recall that Decoupled-RL-BFGS and Decoupled-GP-BFGS correspond 
    to the methods proposed in \citet{petersen2021deep} and \citet{koza1994genetic}, respectively.
    \ALGNAME achieves the best average reward on the test set at the lowest number of function evaluations.}
\label{fig:sr_results}
\end{figure}

\subsubsection{Problem formulation.}
Symbolic regression (SR) \citep{koza1994genetic, bongard2007automated, petersen2021deep,landajuela2021improving,10586218} 
is a classical discrete-continuous optimization problem
with applications in many fields, including robotics, control, and machine learning.
In SR, we have
$\mathcal{ L} = \{
x_1, \ldots, x_d, +, -, \times, \div, \sin, \cos, \ldots\}$ and    
$\hat{\mathcal{L}} = \{\text{const}(\beta)\}$, 
where $\text{const}(\beta)$ represents a constant with value $\beta$.
The design space is a subset of the space of continuous functions, $\mathbb{T} \subset C(V^ \mathbb{R})$, where $V \subset \mathbb{R}^d$ is the function support that depends
on $\mathcal{L}$. The evaluation operator ${\tt eval}$ returns the function which expression tree has the sequence $\tau$ as pre-order traversal (depth-first and then left-to-right).
For example, 
${\tt eval}(\langle +, \cos, y, \times, \text{const}(3.14), \sin, x \rangle) = \cos(y) + 3.14 \times \sin(x)$.
Given a dataset $D = \{(x_1^{(i)}, \dots, x_d^{(i)}, y^{(i)})\}_{i=1}^N$,
the reward function is defined as the inverse of the normalized mean squared error (NMSE) between
$y^{(i)}$ and ${\tt eval}(\tau)(x_1^{(i)}, \dots, x_d^{(i)}), \forall i \in \{1, \dots, N\}$, computed as $\frac{1}{1 + \text{NMSE}}$.
SR has been shown to be NP-hard even for low-dimensional data \citep{virgolin2022symbolic}.
Prefix-dependent positional constraints are given in the appendix.

\subsubsection{Evaluation.}
A key evaluation metric for symbolic regression is the 
\textit{parsimony} of the discovered equations, i.e., 
the balance between the complexity of the identified equations and their ability to fit the data.
A natural way to measure it is to consider the
generalization performance over a test set.
A SR method could find symbolic expressions that overfit the training data
(using for instance overly complex expressions),
but those expressions will not generalize well to unseen data. 
For evaluating the generalization performance of various baselines, we rely on the benchmark datasets detailed in  \Cref{tab:sr_benchmark} of the appendix.

\subsubsection{Results.}
Results in \Cref{fig:sr_results} demonstrate the superior efficiency and generalization capability of \ALGNAME
in the SR setting. In particular, \ALGNAME achieves the best average reward on the test set and the lowest number of function evaluations.
Note that for \ALGNAME we have perfect control over the number of function evaluations as it is 
determined by the number of samples ($10^6$ in this case).
The Decoupled-GP methods exhibit a strong tendency to overfit to the training data and perform poorly on the test set.
This phenomenon is known as the \textit{bloat} problem in the SR literature \citep{silva2009dynamic}.
We observe that the joint optimization of \ALGNAME
avoids this problem and achieve the best generalization performance.

\subsubsection{Literature comparisons.}
In \Cref{tab:disco-dso-comparisons-literature}, we compare \ALGNAME against state-of-the-art methods in the SR literature.
In addition to the baselines \citep{petersen2021deep, koza1994genetic} described above,
we compare against the methods proposed in \citet{biggio2021neural} and \citet{kamienny2022end}.
Since the method in \citet{biggio2021neural} is only applicable to $\leq 3$ dimensions,
we consider the subset of benchmarks with $\leq 3$ dimensions.
We observe that \ALGNAME dominates all baselines in terms of average reward on the full test set.
For the subset of benchmarks with $\leq 3$ dimensions, \ALGNAME achieves comparative performance to
the specialized method in \citet{biggio2021neural}.

\begin{table}[h!]
    \begin{center}
    \scalebox{0.9}{
    \begin{tabular}{l|c|c}
        {Algorithm} & Dim $\leq 3$ & Dim $\geq 1$  \\
        \cline{2-3}
         \hline
         DisCo-DSO & 0.6632 $\pm$ 0.3194 & $\boldsymbol{0.7045}$ $\pm$ 0.3007 \\
         Decoupled-RL-BFGS$^\star$ & 0.6020 $\pm$ 0.4169 & 0.6400 $\pm$ 0.3684 \\
         Decoupled-RL-evo & 0.0324 $\pm$ 0.1095 & 0.0969 $\pm$ 0.2223 \\
         Decoupled-RL-anneal & 0.1173 $\pm$ 0.2745 & 0.1436 $\pm$ 0.3015 \\
         Decoupled-GP-BFGS$^{\star\star}$ & 0.5372 $\pm$ 0.4386 & 0.4953 $\pm$ 0.4344 \\
         Decoupled-GP-evo & 0.0988 $\pm$ 0.1975 & 0.0747 $\pm$ 0.1763 \\
         Decoupled-GP-anneal & 0.1615 $\pm$ 0.2765 & 0.1364 $\pm$ 0.2608 \\
         \citet{kamienny2022end}  & 0.6068 $\pm$ 0.1650 & 0.5699 $\pm$ 0.1065 \\
         \citet{biggio2021neural} & $\boldsymbol{0.6858}$ $\pm$ 0.1995 & N/A\\
    \end{tabular}
    }
    \end{center}
    \caption{Comparison of \ALGNAME against 
    decoupled baselines and the methods proposed in 
    \citet{biggio2021neural} and \citet{kamienny2022end}
    on the symbolic regression task.
    Values are mean $\pm$ standard deviation of the reward across benchmarks 
    (provided in  \Cref{tab:sr_benchmark} of the appendix).
    We group benchmarks because \citet{biggio2021neural} is only applicable to $\leq 3$ dimensions.
    $^\star$\ \citet{petersen2021deep}. $^{\star \star}$\ \citet{koza1994genetic}.
    }
    \label{tab:disco-dso-comparisons-literature}
\end{table} 

\section{Conclusion}

We proposed \ALGNAME (Discrete-Continuous Deep Symbolic Optimization), 
a novel approach to optimization in hybrid discrete-continuous spaces.
\ALGNAME uses a generative model to learn a joint distribution on discrete and continuous design variables
to sample new hybrid designs.
In contrast to standard decoupled approaches,
in which the discrete skeleton is sampled first, and then the continuous variables are optimized separately,
our joint optimization approach samples both discrete and continuous variables simultaneously.
This leads to more efficient use of objective function evaluations,
as the discrete and continuous dimensions of the design space can ``communicate" with each other
and guide the search.
We have demonstrated the benefits of \ALGNAME in challenging problems in symbolic regression and decision tree optimization,
where, in particular, \ALGNAME outperforms the state-of-the-art on univariate decision tree policy optimization for RL.

Regarding the limitations of \ALGNAME, it is important to note that the method relies on domain-specific information to define the ranges of continuous variables. In cases where this information is not available and estimates are necessary, the performance of \ALGNAME could be impacted. Furthermore, in our RL experiments, we constrain the search space to univariate decision trees. Exploring more complex search spaces, such as multivariate or ``oblique" decision trees, remains an avenue for future research. 
\section*{Acknowledgments}
This manuscript has been authored by Lawrence Livermore National Security, LLC under Contract No. DE-AC52-07NA2
7344 with the US. Department of Energy. The United States Government retains, and the publisher, by accepting the
article for publication, acknowledges that the United States Government retains a non-exclusive, paid-up, irrevocable,
world-wide license to publish or reproduce the published form of this manuscript, or allow others to do so, for United
States Government purposes. 
We thank Livermore Computing and the Laboratory Directed Research and Development program (21-SI-001) for their support.
Release code is LLNL-CONF-854776.

\bibliography{discrete_optimization,citation_rl}
\bibliographystyle{plainnat}

\newpage
\appendix
\onecolumn

\section{Pseudocode for \ALGNAME}
\label{app:long-pseudocode}

In this section we present pseudocode for the \ALGNAME algorithm. 
The algorithm is presented in \Cref{alg:discodso}.
We also provide pseudocode for the discrete-continuous sampling procedure in \Cref{alg:sampling}. 
\Cref{alg:sampling} is called multiple times to form a batch in \Cref{alg:discodso}. 
Note that for decision tree policies for reinforcement learning,
the distribution, $\mathcal{D}$, used for sampling the next continuous token in \Cref{alg:sampling} is only truncated with bounds produced with \Cref{alg:decision-tree-parameter-bounds} at a decision tree node. 
Otherwise, the distribution is unbounded.

\def\lbfull{(l, \beta)_{1:{T_i}}^{(i)}}
\def\lbfullt{\tau^{(i)}}
\def\lb{(l, \beta)_{1:T}^{(i)}}

\begin{algorithm*}[h!]
\caption{Discrete-Continuous Deep Symbolic Optimization}
\label{alg:discodso}
\begin{algorithmic}[1]
\INPUT batch size $N$, reward function $R$, risk factor $\epsilon$, policy gradient function $S$, entropy coefficient $\lambda_\mathcal{H}$, learning rate $\alpha$
\OUTPUT Best fitting design $\tau^\star = \langle (l_1, \beta_1), \ldots, (l_T, \beta_T) \rangle^\star $
\STATE Initialize \NN\ with parameters $\theta$, defining distribution over designs $p(\cdot | \theta)$
\STATE \textbf{repeat}
\begin{ALC@rpt}
\STATE $\mathcal{T} \leftarrow \{ \tau^{(i)} = \lbfull \sim p( \cdot | \theta)\}_{i=1}^N$ \RCOMMENT{Sample batch of $N$ discrete-continous designs (Algorithm \ref{alg:sampling})}
\STATE $\mathcal{R} \leftarrow \{R(\lbfullt) \}_{i=1}^N$ \RCOMMENT{Compute rewards}
\STATE $R_\eps \leftarrow (1 - \eps)$-quantile of $\mathcal{R}$ \RCOMMENT{Compute reward threshold}
\STATE $\mathcal{T} \leftarrow \{\lbfullt : R(\lbfullt) \geq R_\eps \}$ \RCOMMENT{Select subset of expressions above threshold}
\STATE $\mathcal{R} \leftarrow \{R(\lbfullt) : R(\lbfullt) \geq R_\eps \}$ \RCOMMENT{Select corresponding subset of rewards}
\STATE $\hat{g_1} \leftarrow \textrm{ReduceMean}( (\mathcal{R} - R_\eps) S(\mathcal{T},\theta ))$ \RCOMMENT{Compute risk-seeking policy gradient}
\STATE $\hat{g_2} \leftarrow \textrm{ReduceMean}( -\lambda_\mathcal{H}\nabla_\theta \mathcal{H}(\mathcal{T} | \theta) )$ \RCOMMENT{Compute entropy gradient}
\STATE $\theta \leftarrow \theta + \alpha (\hat{g_1} + \hat{g_2})$ \RCOMMENT{Apply gradients}
\SHORTIF{$\max\mathcal{R} > R(\tau^\star)$} {$\tau^\star \leftarrow \tau^{(\arg\max \mathcal{R})}$} \RCOMMENT{Update best discrete-continuous design}
\end{ALC@rpt}
\STATE \textbf{return} $\tau^\star$
\end{algorithmic}
\end{algorithm*}

\begin{algorithm*}[h!]
\caption{Discrete-continuous sampling}
\label{alg:sampling}
\begin{algorithmic}[1]
\INPUT parameters of the given \NN\ $\theta$,  token library $\mathcal{L}$ 
\OUTPUT sequence $\tau = \langle (l_1, \beta_1), \ldots, (l_T, \beta_T) \rangle$ sampled from the \NN
\STATE $\tau = \tau_{1:0}  \leftarrow [\cdot]$ \RCOMMENT{Initialize empty sequence}
\STATE \textbf{for} $i = 1,2,\hdots$ \textbf{do}
\begin{ALC@rpt}
    \STATE $(\psi^{(i)}, \phi^{(i)}) \leftarrow \NN(\tau_{1:(i-1)} = (l,\beta)_{1:(i-1)};\theta)$ \RCOMMENT{Emit two outputs for each token}
    \STATE Compute $\mathcal{C}_{\tau_{1:(i-1)}} \subseteq \mathcal{L}$ \RCOMMENT{Compute unfeasible tokens}
    \STATE $\psi^{(i)}_{\mathcal{C}_{\tau_{1:(i-1)}}} \leftarrow -\infty$ \RCOMMENT{Set unfeasible tokens to $-\infty$}
    \STATE $l_i\leftarrow $ Categorical($\psi^{(i)}$) \RCOMMENT{Sample the next discrete token}
    \IF{$l_i \in \bar{\mathcal{L}}$}
        \STATE $\beta_i \leftarrow \mathcal{U}_{[0,1]}(\cdot)$ \RCOMMENT{Sample the next continuous token}
    \ELSIF{$l_i \in \hat{\mathcal{L}}$} 
        \STATE{$\beta_i \leftarrow \mathcal{D}(\cdot | l_i, \phi^{(i)})$} \RCOMMENT{Sample the next continuous token}
    \ENDIF
    \STATE $\tau_i \leftarrow (l_i,\beta_i)$ \RCOMMENT{Joint discrete and continuous token}
    \STATE $\tau \leftarrow \tau\mathbin\Vert\tau_i$ \RCOMMENT{Append token to traversal}
\end{ALC@rpt}
\end{algorithmic}
\end{algorithm*}

\section{Additional algorithm details}
\label{app:rspg}

\subsection{Risk-seeking policy gradient for hybrid discrete-continuous action space}
The derivation of the risk-seeking policy gradient 
for the hybrid discrete-continuous action space 
follows closely the derivation in 
\citet{petersen2021deep} (see also \citet{tamar2014policy}).
The risk-seeking policy gradient for a univariate sequence $\tau$ is given by
\begin{align*}
    \nabla_{\theta} J_{\varepsilon}(\theta) &= \mathbb{E}_{\tau \sim p(\tau | \theta)} 
    \left[ \left( R(\tau) - R_{\varepsilon}(\theta) \right) \nabla_{\theta} \log p(\tau | \theta) \mid R(\tau) \geq R_{\varepsilon}(\theta) \right].
\end{align*}
In the hybrid discrete-continuous action space case,
we have
$\tau = \langle (l_1, \beta_1), \ldots, (l_T, \beta_T)\rangle$ and
\begin{align}
p(\tau | \theta) = 
\prod_{i=1}^{|\tau|} p(\tau_i | \tau_{1:(i-1)}, \theta) =
\prod_{i=1}^{|\tau|} p((l_i,\beta_i) | (l,\beta)_{1:(i-1)}, \theta).
\label{eq:chain_rule}
\end{align}
Thus, using the convenient notation $A(\tau, \varepsilon, \theta) = R(\tau) - R_{\varepsilon}(\theta)$,
the risk-seeking policy gradient for the hybrid discrete-continuous action space is given by
\begin{align*}
    & \nabla_{\theta} J_{\varepsilon}(\theta) = \mathbb{E}_{\tau \sim p(\tau | \theta)}  
            \left[ A(\tau, \varepsilon, \theta) \nabla_{\theta} \log p(\tau | \theta) \mid A(\tau, \varepsilon, \theta) \geq 0 \right] \\
    &= \mathbb{E}_{\tau \sim p(\tau | \theta)} \left[ A(\tau, \varepsilon, \theta) \nabla_{\theta} \log \prod _{i=1}^{|\tau|} p((l_i, \beta_i) | (l,\beta)_{1:(i-1)}, \theta) \mid A(\tau, \varepsilon, \theta) \geq 0 \right] \\
    &= \mathbb{E}_{\tau \sim p(\tau | \theta)} \left[ A(\tau, \varepsilon, \theta) \sum_{i=1}^{|\tau|} \nabla_{\theta} \log p((l_i, \beta_i) | (l,\beta)_{1:(i-1)}, \theta) \mid A(\tau, \varepsilon, \theta) \geq 0 \right] \\
    &= \mathbb{E}_{\tau \sim p(\tau | \theta)} \left[ A(\tau, \varepsilon, \theta) \sum_{i=1}^{|\tau|} 
            \begin{cases}
            \nabla_{\theta} \log p(l_i | (l,\beta)_{1:(i-1)}, \theta), & \text{if } l_i \in \bar{\Lcal} \\
            \nabla_\theta \log p(\beta_i | l_i, (l,\beta)_{1:(i-1)}, \theta) 
            + \nabla_\theta \log p(l_i | (l,\beta)_{1:(i-1)}, \theta), & \text{if } l_i \in \hat{\Lcal}
            \end{cases}
            \mid A(\tau, \varepsilon, \theta) \geq 0 \right].
    \end{align*}

In practice, we use the following estimator for the risk-seeking policy gradient:
\begin{align*}
&\nabla_{\theta} J_{\varepsilon}(\theta)  \approx 
        \frac{1}{M} \sum_{i=1}^{M} \tilde{A}(\tau^{(i)}, \varepsilon, \theta) \sum_{j=1}^{|\tau^{(i)}|} 
        \begin{cases}
        \nabla_{\theta }\log p(l_j^{(i)} | (l,\beta)_{1:(j-1)}^{(i)}, \theta), & \text{if } l_j^{(i)} \in \bar{\Lcal} \\
        \nabla_\theta \log p(\beta_j^{(i)} | l_j^{(i)}, (l,\beta)_{1:(j-1)}^{(i)}, \theta)
        + \\ \quad \quad \nabla_\theta \log p(l_j^{(i)} | (l,\beta)_{1:(j-1)}^{(i)}, \theta), & \text{if } l_j^{(i)} \in \hat{\Lcal}
        \end{cases}\\
        & \ \ \ \ \ \ \ \ \ \ \ \ \ \ \ \ \ \ \ \ \ \ \ \ \ \ \ \ \ \ \ \ \ \ \ \ \ \ \ \ \ \ \ \ \ \ \ \ \ \ \ \ \ \ \ 
        \ \ \ \ \ \ \ \ \ \ \ \ \ \ \
        \cdot \mathbb{I}(\tilde{A}(\tau^{(i)}, \varepsilon, \theta) \geq 0),
\end{align*}
where $\tau^{(i)} = \langle (l_1^{(i)}, \beta_1^{(i)}), \ldots, (l_T^{(i)}, \beta_T^{(i)})\rangle$
is the $i$-th trajectory sampled from $p(\tau | \theta)$,
$M$ is the number of trajectories used in the estimator,
and $A(\tau^{(i)}, \varepsilon, \theta) \approx \tilde{A}(\tau^{(i)}, \varepsilon, \theta) = R(\tau^{(i)}) - \tilde{R}_{\varepsilon}(\theta)$,
with $\tilde{R}_{\varepsilon}(\theta)$ being an estimate of $R_{\varepsilon}(\theta)$.

\subsection{Entropy derivation in the hybrid discrete-continuous action space}
As in \citet{petersen2021deep}, we add entropy to the loss function as a bonus. Since there is a continuous component in the library, the entropy for the distribution of $\tau_i$ in the sequence is
\begin{equation*}
\mathcal{H}_i  
= \sum_{\distok\in\paramtokset}p(\distok | (l, \beta)_{1:(i-1)},\theta)\mathcal{H}_{\contok\sim\mathcal{D}(\contok|\distok, \phi_\distok)}(\contok|\distok)  
+ \mathcal{H}_{\distok\sim p(\distok | (l, \beta)_{1:(i-1)},\theta)}(\distok).
\end{equation*}
For the derivation, recall that, for a distribution $\mathcal{D}$, the entropy is defined as
$
    \mathcal{H}(\mathcal{D}) = -\int_{-\infty}^{\infty} \mathcal{D}(x) \log \mathcal{D}(x)\,dx,
$
and that, in \ALGNAME, we add an entropy regularization term for each
distribution $p((\distok_i, \contok_i) | (\distok, \contok)_{1:(i-1)},\theta)$
encountered during the rollout. 
Thus, we have
\begin{equation*}
\begin{split}
\mathcal{H}_i  = 
& - \sum_{\distok_i\in\paramtokset} \int_{-\infty}^{\infty} p((\distok_i, \contok_i) | (\distok, \contok)_{1:(i-1)},\theta) \log p((\distok_i, \contok_i) | (\distok, \contok)_{1:(i-1)},\theta)\,d\contok_i \\
& \hspace{4cm} - \sum_{\distok_i\in\distokset} p(\distok_i| (\distok, \contok)_{1:(i-1)},\theta) \log p(\distok_i | (\distok, \contok)_{1:(i-1)},\theta)\\
  =  & - \sum_{\distok_i\in\paramtokset} \int_{-\infty}^{\infty} p(\contok_i | \distok_i) p(\distok) \log \left( p(\contok_i | \distok_i) p(\distok_i) \right) \,d\contok_i 
    - \sum_{\distok_i\in\distokset} p(\distok_i) \log p(\distok_i) \\
  = & - \sum_{\distok_i\in\paramtokset} \int_{-\infty}^{\infty} p(\contok_i | \distok_i) p(\distok_i) \log p(\contok_i | \distok_i)\,d\contok_i - \sum_{\distok_i\in\paramtokset} \int_{-\infty}^{\infty} p(\contok_i | \distok_i) p(\distok_i) \log p(\distok_i)\,d\contok_i \\
  & \ \ \ \ \ \ \ \  - \sum_{\distok_i\in\distokset} p(\distok_i) \log p(\distok_i) \\
  = & - \sum_{\distok_i\in\paramtokset} p(\distok_i) \int_{-\infty}^{\infty} p(\contok_i | \distok_i) \log p(\contok_i | \distok_i)\,d\contok_i 
  - \sum_{\distok_i\in\paramtokset}  p(\distok_i) \log p(\distok_i) \int_{-\infty}^{\infty} p(\contok_i | \distok_i) \,d\contok_i \\
  & \ \ \ \ \ \ \ \  - \sum_{\distok_i\in\distokset} p(\distok_i) \log p(\distok_i) \\
   = & \sum_{\distok_i\in\paramtokset}p(\distok_i | (\distok, \contok)_{1:(i-1)},\theta)\mathcal{H}_{\contok_i\sim\mathcal{D}(\contok_i|\distok_i, \phi_{\distok_i})}(\contok_i|\distok_i)  \\ 
   & \ \ \ \ \ \ \ \ - \sum_{\distok_i\in\mathcal{L}}p(\distok_i | (\distok, \contok)_{1:(i-1)},\theta) \log p(\distok_i | (\distok, \contok)_{1:(i-1)},\theta).
\end{split}
\end{equation*}
Note that we have removed the conditioning elements $\theta$ and $(\distok, \contok)_{1:(i-1)}$ 
in some terms in the above derivation for brevity.


\section{Additional experimental results per task}
\label{app:exp}

\subsection{Parameterized bitstring task}
\label{app:bitstring_error_functions}

\subsubsection{Objective functions.}
\begin{figure}[h]
    \centering
    \subfigure[$f_1$]{\includegraphics[width=0.48\linewidth]{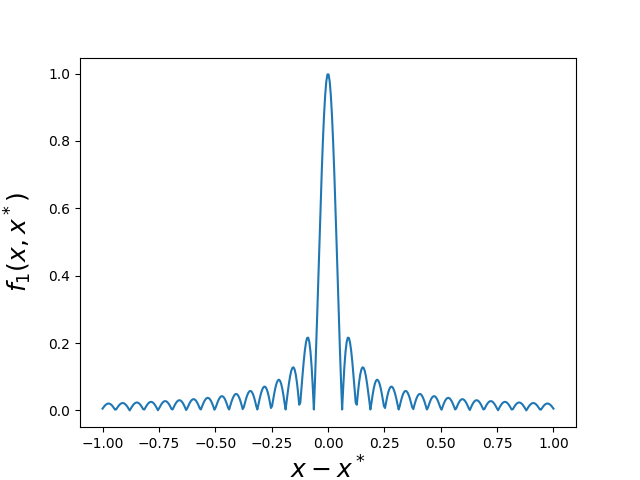}} \hfil
    \subfigure[$f_2$]{\includegraphics[width=0.48\linewidth]{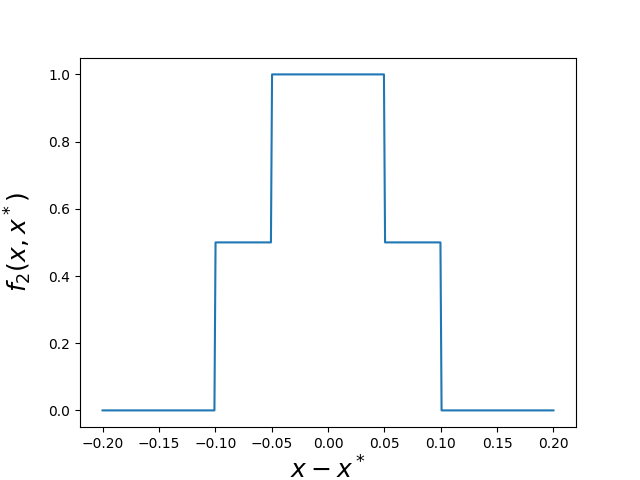}}
    \caption{Objective functions in \eqref{eq:bitstring_error_functions} against the difference $x-x^*$.}
    \label{fig:bitstring_error_functions}
\end{figure}

The objective functions $f_1$ and $f_2$ in \Cref{eq:bitstring_error_functions} are plotted in Figure~\ref{fig:bitstring_error_functions}. They are both non-differentiable and difficult to be optimized by Quasi-Newton methods.


\subsubsection{Objective gap between the best solutions found by \ALGNAME and baselines for the parameterized bitstring task.}
In Table~\ref{tab:bitstring_gap}, we show the gap between the best solutions obtained 
by \ALGNAME and the baselines for the parameterized bitstring task. The gap is computed by taking the best \ALGNAME solutions and the best solutions of the baselines for each seed. The reward differences are averaged over the 5 seeds.

\begin{table}[h!]
    \begin{center}
    \begin{tabular}{c|c|c|c|c}
    Baseline & $f_1$, $\alpha = 0.5$ & $f_1$, $\alpha = 0.9$ & $f_2$, $\alpha = 0.5$ & $f_2$, $\alpha = 0.9$ \\
    \hline
    Decoupled-RL-BFGS & 0.1263 & 0.0123 & 0.1433 & 0.0253 \\
    Decoupled-RL-evo & 0.0970 & 0.0100 & 0.0833 & 0.0187 \\
    Decoupled-RL-anneal & 0.1211 & 0.0095 & 0.1000 & 0.0140 \\
    \end{tabular}
    \end{center}
    \caption{Gap between the best solutions found by \ALGNAME and baselines for the parameterized bitstring task. Positive values indicate that \ALGNAME found better solutions.}
    \label{tab:bitstring_gap}
\end{table} 
\subsection{Decision tree policies for reinforcement learning}
\label{app:control}

\subsubsection{Results for MountainCar-v0 and CartPole-v1 environments.}
\begin{figure}[h]
    \centering
    \subfigure[MountainCar-v0]{\includegraphics[width=0.48\linewidth]{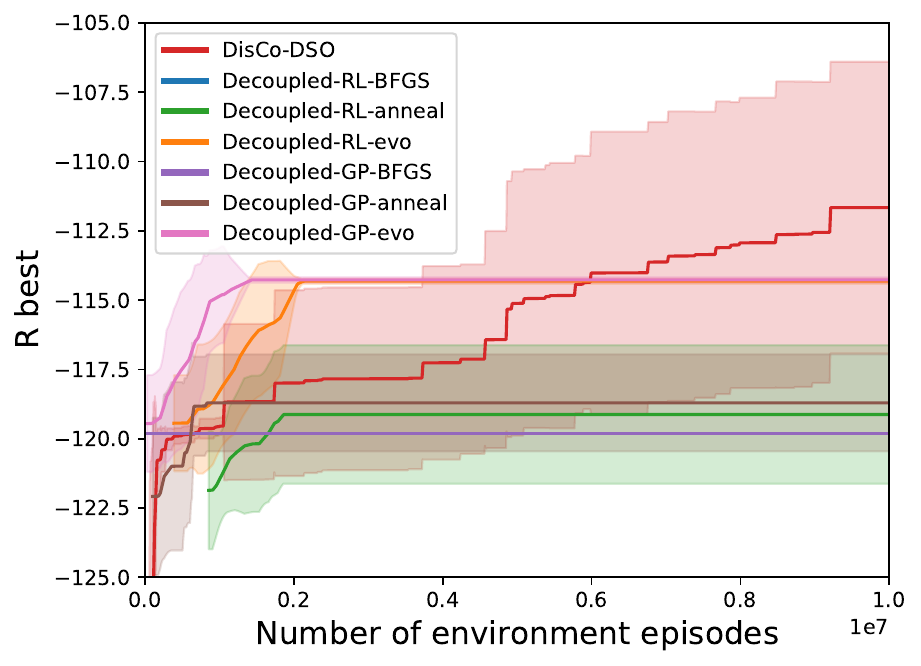}}
    \hfil
    \subfigure[CartPole-v1]{\includegraphics[width=0.48\linewidth]{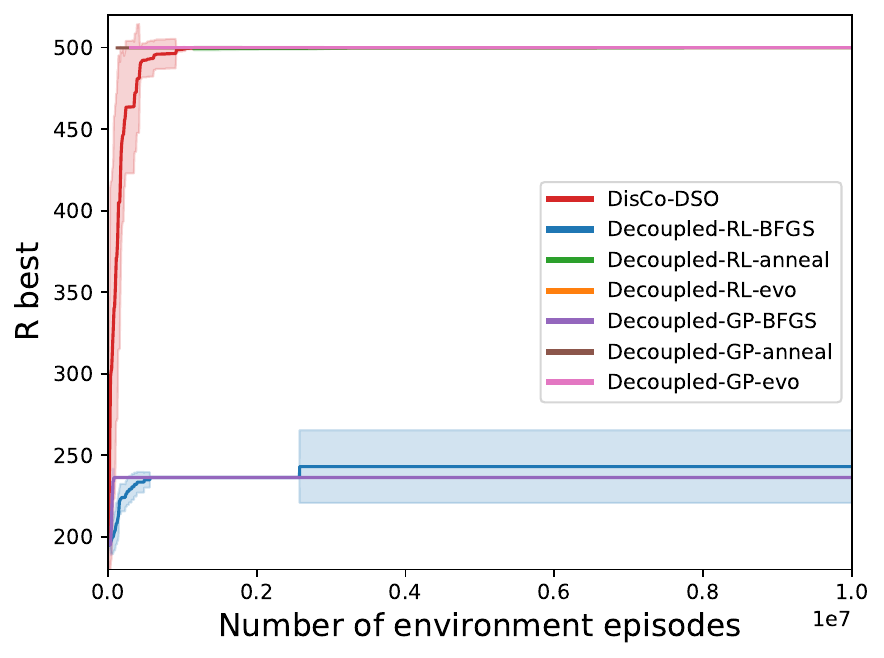}}
    \caption{Reward of the best solution versus number of function evaluations on the \DT policy task, for MountainCar-v0 and CartPole-v1.}
    \label{fig:learning_curves_mc_cp}
\end{figure}

\begin{figure}[t]
\centering
\subfigure[MountainCar-v0]{\includegraphics[width=0.48\linewidth]{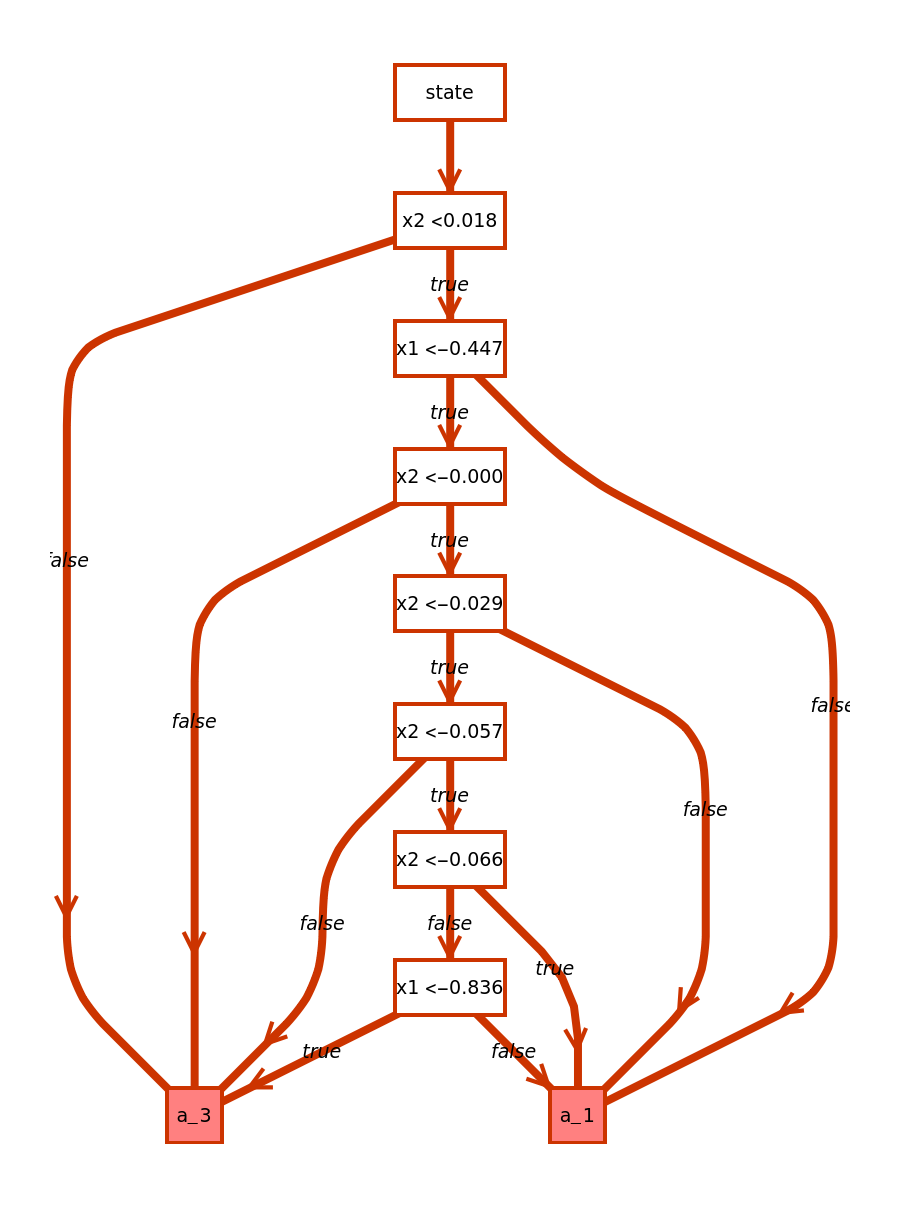}} \hfil
\subfigure[CartPole-v1]{\includegraphics[width=0.48\linewidth]{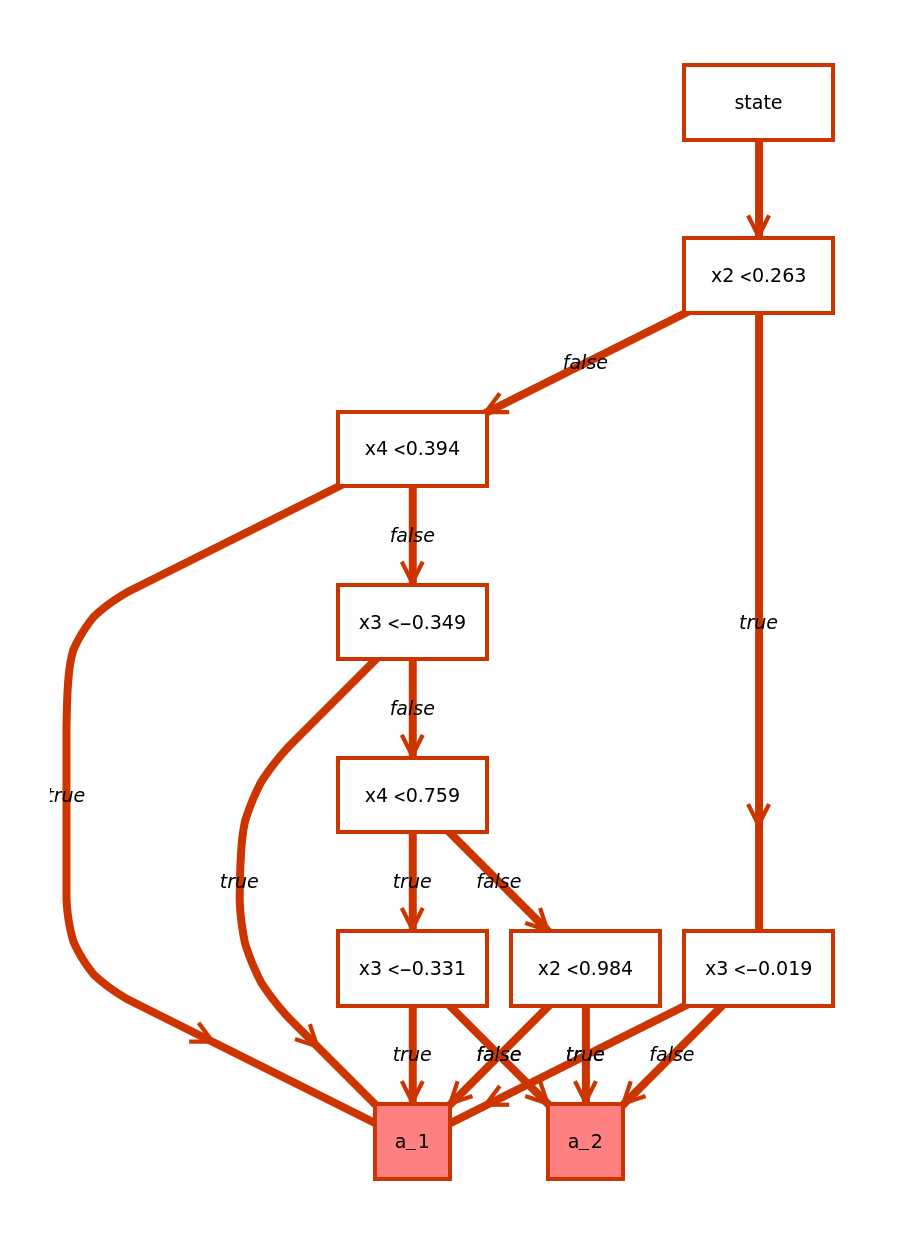}}
\caption{Best \DTs found by \ALGNAME on the \DT policy tasks for MountainCar-v0 and CartPole-v1.}
\label{fig:best_decision_trees_mc_cp}
\end{figure}

In this section, we provide results for the MountainCar-v0 and CartPole-v1 environments.
In \Cref{fig:learning_curves_mc_cp}, we show 
the best reward versus number of environment episodes
for MountainCar-v0 and CartPole-v1.
\Cref{fig:best_decision_trees_mc_cp} shows the best decision trees found by \ALGNAME for MountainCar-v0 and CartPole-v1.

\subsubsection{Note on ``oblique" decision trees.}
\cite{custode2023evolutionary} consider multivariate, or "oblique" decision trees. 
These are trees where the left-hand side of a decision node is composed of an expression, while the right-hand side is still a boolean decision parameter $\beta_n$. While these trees perform well on more complex environments such as LunarLander-v2 (published results report an average test score of 213.09), we do not compare against them here as the search space is drastically different.



\subsubsection{Objective gap between the best solutions found by \ALGNAME and baselines for the decision tree policy task.}
In \Cref{tab:control_results_gap} and \Cref{tab:control_results_literature_gap}, we show the objective gap between the best solutions found by \ALGNAME and the baselines for the \DT policy task. The gap is computed by taking the best \ALGNAME solutions and the best solutions of the baselines.

\begin{table}[h]
    \begin{center}
    \begin{tabular}{c|c|c|c|c}
        Baseline & CartPole-v1 & MountainCar-v0 & LunarLander-v2 & Acrobot-v1 \\
        \hline
        Decoupled-RL-BFGS & 260.00 & 7.80 & 74.80 & 4.22 \\
        Decoupled-RL-evo & 0.00 & 2.60 & 80.60 & 6.52 \\
        Decoupled-RL-anneal & 0.00 & 7.10 & 79.60 & 4.52 \\
        Decoupled-GP-BFGS & 265.00 & 7.70 & 64.80 & 4.72 \\
        Decoupled-GP-evo & 0.00 & 2.80 & 54.60 & 7.92 \\
        Decoupled-GP-anneal & 0.00 & 6.10 & 43.20 & 4.52 \\
    \end{tabular}
    \end{center}
    \caption{Objective gap between the best solutions found by \ALGNAME and baselines for the \DT policy task. Positive values indicate that \ALGNAME found better solutions.}
    \label{tab:control_results_gap}
\end{table}

\begin{table}[h]
    \begin{center}
    \begin{tabular}{c|c|c|c|c}
        Baseline & CartPole-v1 & MountainCar-v0 & LunarLander-v2 & Acrobot-v1 \\
        \hline
        Evolutionary DTs & 0.42 & 3.96 & 186.86 & 20.54 \\
        Cascading DTs & 3.37 & 99.03 & 326.26 & 5.56 \\
        Interpretable DDTs & 110.21 & 71.24 & 219.62 & 421.28 \\
        Bayesian Optimization & 414.53 & 99.03 & 211.38 & 14.41 \\
    \end{tabular}
    \end{center}
    \caption{Objective gap between the best solutions found by \ALGNAME and baselines from the literature for the \DT policy task. Positive values indicate that \ALGNAME found better solutions.}
    \label{tab:control_results_literature_gap}
\end{table}

\subsection{Symbolic regression for equation discovery with constants}
\label{app:symbolic_regression}

\subsubsection{Additional details on the evaluation procedure for symbolic regression.}
For evaluating the generalization performance of various baselines, we rely on the benchmark datasets detailed in  \Cref{tab:sr_benchmark}. 
The test set is obtained by expanding the benchmark's domain $(a,b)$ and increasing the number of data points on which an expression is evaluated.
Since all experiments were conducted with 10 different random seeds, each random seed leads to a distinct ``best" expression for that specific run. 
We take each of these 10 best expressions and compute 
the reward obtained on the evaluation dataset.
Additionally, we calculate the number of function evaluations required to arrive at each expression.
Subsequently, we aggregate the set of 10 evaluation rewards to calculate a single average reward and a single average number of function evaluations for each dataset.
Following this procedure, we average over all the datasets to get an overall evaluation for each method.

\subsubsection{Benchmarks.}
In \Cref{tab:sr_benchmark}, we provide a compilation of benchmarks used in the symbolic regression task. The list comprises Livermore benchmarks from \cite{mundhenk2021symbolic}, Jin benchmarks from \cite{jin2020bayesian}, and Neat and Korn benchmarks from \cite{trujillo201621}. We also introduce Constant benchmarks, which are a variation of the Nguyen benchmarks from \cite{quang2011} where floating constants are added to increase the complexity of the problem. Note that the selection of these benchmarks is not arbitrary. The linear and non-linear dependency of the expressions on $\beta_i$ is taken into consideration. Non-linear functions involving $\beta_i$ exhibit more complicated expressions, thus rendering more challenges for the optimization problem. On the other hand, linear functions, such as Jin-1, employ constants as coefficients for each variable term, thereby simplifying the optimization problems into linear regression problems. Consequently, we have selected approximately 25 non-linear functions and 20 linear functions, resulting in a total of 45 benchmark datasets. Exploring whether joint discrete-continuous optimization can outperform other classes of methods for both non-linear functions and linear functions can be a focus for future research. 

\subsubsection{Evaluations sets.}
To create the evaluation set, we adhere to a straightforward rule: we take the training set of each benchmark function, double the size of its domain, and double the number of points at which it is computed. For instance, a benchmark function with a training domain of $(-1, 1)$ and 20 points in that domain would have an evaluation set spanning $(-2, 2)$ with 40 data points within the expanded domain.

\begin{table}[h!]
    \tiny
    \begin{center}
    \begin{tabular}{c|c|c}
        Benchmark Name & Expression & Dataset \\ \hline
        Livermore2-Vars2-2 & $x_1\,x_2\,\left(x_1 + x_1\,\sqrt{x_2\,(\beta_1 + x_1 + \beta_2\,x_1\,x_2 + \beta_3\,x_2^3)}\right)$ & U(-10,10,1000) \\ \hline
        Livermore2-Vars2-4 & $\beta_1 \,x_1+\beta_2 \,x_2+(x_1-x_2)^2$ & U(-10,10,1000) \\ \hline
        Livermore2-Vars2-6 & $-\left((x_1 - x_1^2 - x_2)\,x_2\right) + \frac{(\beta_1\,x_1)}{(\beta_2 + \beta_3\,x_2)}$ & U(-10,10,1000) \\ \hline
        Livermore2-Vars2-7 & $1 + \beta_1\,\sqrt{x_1} + \beta_2\,x_1 + \frac{(\beta_3\,x_2)}{x_1} + \beta_4\,\sqrt{-x_1 + x_2}$ & U(-10,10,1000) \\ \hline
        Livermore2-Vars2-8 & $\beta_1\,x_1^2 + \beta_2\,x_2^2 + \beta_3\,x_2^3 + \frac{((\sqrt{x_1} - x_1)\,\log{(x_2)})}{\log{(x_1)}}$ & U(-10,10,1000) \\ \hline
        Livermore2-Vars2-12 & $\sqrt{(x_1 + \frac{x_2}{e^{x_2}})\,\log(\beta_1\,x_1^3 + \beta_2\,x_1^2\,x_2 + \beta_3\,x_2^3)}$ & U(-10,10,1000)\\ \hline
        Livermore2-Vars2-24 & $\beta_1 + \beta_2\,x_1 + \beta_3\,(x_1 - x_2^2)^2$ & U(-10,10,1000) \\ \hline
        Livermore2-Vars3-4 & $\beta_1+\beta_2 x_2 x_3+x_1+x_2+\sqrt{\cos (x_2})+x_3$ & U(-10,10,1000) \\ \hline
        Livermore2-Vars2-9 & $\beta_1 + x_1 + x_1^2 - x_2 + \beta_2\,\sqrt{1 + \beta_3\,x_1 + \beta_4\,x_2}$ & U(-10,10,1000)\\ \hline
        Livermore2-Vars2-16 & $-\left((x_1 - x_1^2 - x_2)\,x_2\right) + \frac{(\beta_1\,x_1)}{(\beta_2 + \beta_3\,x_2)}$ & U(-10,10,1000) \\ \hline
        Livermore2-Vars2-17 & $x_1 - \sqrt{x_1}\,x_2 - \sin{\left(\log{(\beta_1\,x_1 + \beta_2\,x_1^3 + \beta_3\,x_2^2)}\right)}^2$ & U(-10,10,1000) \\ \hline
        Livermore2-Vars2-19 & $\left(x_1 + \frac{(\beta_1*\cos{(x1)})}{\sqrt{1 + \beta_2\,x_1 + \beta_3\,x_1\,(x_2^2 + \log{(x_2)})}}\right)\,(x_1 + \sin{(1)})$ & U(-10,10,1000)\\ \hline
        Livermore2-Vars2-22 & $\frac{\log{(\beta_1\,(-1 + \beta_2\,x_1)^2 + \frac{(e^{x_2}\,x_1)}{\sqrt{\cos{(\beta_3*x_2)}}})}}{\sqrt{x_1}}$ & U(-10,10,1000)\\ \hline
        Livermore2-Vars2-23 & $x_1^2 + \beta_1\,\sqrt{1 + \frac{(\beta_2\,(\beta_3 - x_2)\,(\beta_4 + \log{x_2}))}{x_1}}$ & U(-10,10,1000) \\ \hline
        Livermore2-Vars3-2 & $\frac{(\beta_1\,(\beta_2\,x_1 - x_2)^2\,(\beta_4 - x_3)^2)}{(1 + \beta_3\,x_2^2)^2}$ & U(-10,10,1000) \\ \hline
        Livermore2-Vars3-8 & $\frac{x_1}{\left((e^{(x_2 + x_2^4)} + \beta_1\,x_1^2 + \beta_2\,x_1^2\,x_3 + \beta_3\,x_3^2)\,(-x_2 + x_3 + \sqrt{\frac{\log{(x_3)}}{(x_1^2 + \sqrt{x_3})}})\right)}$ & U(-10,10,1000) \\ \hline
        Livermore2-Vars3-9 & $x_1\,\sin{\left(\frac{(\beta_1\,x_1)}{\sqrt{\beta_2\,x_1^2 - \frac{(x_1\,(-1 + x_1 + x_2^2)^4\,\sqrt{x_2 + \beta_3\,x_1\,x_3^2})}{x_3}}}\right)}$ & U(-10,10,1000) \\ \hline
        Livermore2-Vars3-11 & $\beta_1\,x_1 + \frac{x_1}{x_2} + \beta_2\,x_2 + \beta_3\,x_3$ & U(-10,10,1000) \\ \hline
        Livermore2-Vars3-12 & $\beta_1+\beta_2 \,x_1 x_2^2+x_2+x_3$ & \\ \hline
        Livermore2-Vars3-17 & $\beta_1 \,x_1+\beta_2 \,x_2+x_1 \sqrt{x_2} \cos (x_1)+x_3+1$ & U(-10,10,1000) \\ \hline
        Livermore2-Vars3-20 & $\beta_1+\beta_2 \sqrt{ \frac{\sqrt{x_2}}{\sqrt{x_3}} }+x_1$ &  U(-10,10,1000) \\ \hline
        Livermore2-Vars3-24 & $\beta_1+\beta_2 \,x_1 \,x_2^2+x_2+x_3$ & U(-10,10,1000) \\ \hline
        Livermore2-Vars4-8 & $-x_1 + x_1\,\left(x_1 + x_4 + \sin{\left(\frac{(-(e^{e^{x_3}}\,x_1) + x_2)}{(\beta_1\,x_1^2\,x_3 + \beta_2\,x_3^2 + \beta_3\,x_3^3)}\right)}\right)$ & U(-10,10,1000) \\ \hline
        Livermore2-Vars4-16 & $x_3 \left(\frac{\beta_1}{x_3}-x_4\right)+\frac{\beta_2 \,x_4}{x_1}+\sqrt{x_2 \left(x_1^2 \left(-e^{x_2}\right)-x_2\right)}$ & U(-10,10,1000) \\ \hline
        Livermore2-Vars4-18 & $x_1 + \sin{\left(2\,x_2 + x_3 + \beta_1\,\sqrt{\beta_2 + \beta_3\,x_2^3 + x_2\,x_3^2} - e^{x_1}\,x_4 + \log{((-x_1 + x_2)\,\log{(x_2)})}\right)}$ & U(-10,10,1000) \\ \hline
        Livermore2-Vars4-23 & $-(\frac{x_1}{x_2}) + x_2 + \beta_1\,x_2\,x_3 + \frac{(\beta_2\,x_3)}{\sqrt{x_4}} + \beta_3\,\sqrt{x_4} + \log{(x_1)}$ & U(-10,10,1000) \\ \hline
        Livermore2-Vars6-22 & $x_1 + \cos{(x_2 + \beta_1\,x_5)}\,(x_4 - \cos{(x_6)}\,\sin{\left(\beta_2\,(\beta_3 + x_3 - x_4)\right)})$ & U(-10,10,1000) \\ \hline
        Livermore2-Vars6-23 & $x_1 + x_4 + \log{\left(x_1^2 + x_1\left(\beta_1\,\sqrt{\beta_2\,x_1 + x_2 + \beta_3\,(x_3 - \frac{x_4}{x_5})} - x_6\right)\right)}$ & U(-10,10,1000)\\ \hline
        Livermore2-Vars6-24 & $\frac{\left(\beta_1\,\left(\beta_2\left(\beta_3\,x_2 + \frac{\sqrt{x_3}}{(\sqrt{-x_4 + x_5}\,x_6)}\right)^{1/4} + \sin{(x_1)}\right)\right)}{x_6}$ & U(-10,10,1000)\\ \hline
        Livermore2-Vars7-14 & $\beta_1 x_6+\beta_2\, x_1^2 x_6-\cos \left(x_7 \left(\frac{x_2^2 x_5^2 x_6^2 (x_1+x_3 \,x_4 \,x_6)^2+x_6}{x_1}+x_7\right)\right)$ & U(-10,10,1000) \\ \hline
        Livermore2-Vars7-23 & $-\left(\frac{(\sqrt{x_3}x_4)}{(x_3 + x_5 + \beta_1\,x_2\,x_3\,x_5 + \beta_2\,x_2\,x_6 + \beta_3\,x_4\,x_6\,x_7 + (-x_6 + x_7)^2)}\right) + x_1\,\cos{(x_1)} + \cos{(x_2)}$ & U(-10,10,1000) \\ \hline
        Jin-1 & $\beta_1 \,x_1^3+\beta_2 \,x_1^4+\beta_3 \,x_2+\beta_4 \,x_2^2$ & U(-3,3,100) \\ \hline
        Jin-2 & $\beta_1 + \beta_2\,x_1^2 + \beta_3\,x_2^3$ & U(-3,3,100) \\ \hline
        Jin-3 & $\beta_1\,x_1 + \beta_2\,x_1^3 + \beta_3\,x_2 + \beta_4\,x_2^3$ & U(-3,3,100) \\ \hline
        Jin-6 & $\beta_1 \,x_1 \,x_2+\beta_2 \sin ((\beta_3+x_1) (\beta_4+x_2))$ & U(-3,3,100) \\ \hline
        Korn-12 & $2+\beta_1 \cos (\beta_2 \,x_1) \sin (\beta_3 \,x_5)$ & U(-50,50,100) \\ \hline
        Neat-7 & $2+\beta_1 \cos (\beta_2 \,x_1) \sin (\beta_3 \,x_2)$ & U(-50,50,10000) \\ \hline
        Constant-1 & $\beta_1 \,x_1+\beta_2 \,x_1^2+\beta_3 \,x_1^3$ & U(-1,1,20) \\  \hline
        Constant-2 & $\beta_1+\sin \left(x_1^2\right) \cos (x_1)$ & U(-1,1,20) \\ \hline
        Constant-3 & $\cos(x_1\,x_2)\sin(\beta_2\,x_1)$ & U(0,1,20) \\ \hline
        Constant-4 & $\beta_1 \,x_1^{x_2}$ & U(0,1,20) \\ \hline
        Constant-5 & $\beta_1\,\sqrt{x_1}$ & U(0,4,20) \\ \hline
        Constant-6 & $x_1^{\beta_1}$ &  U(0,4,20)\\ \hline
        Constant-7 & $2\cos(x_2)\sin(\beta_1\,x_1)$ & U(0,1,20) \\ \hline
        Constant-8 & $\log (\beta_1+x_1)+\log \left(\beta_2+x_1^2\right)$ & U(0,4,20)
    \end{tabular}
    \end{center}
    \caption{List of benchmarks used for symbolic regression task. Each benchmark includes input variables denoted as $x_1, x_2, \ldots, x_d$, along with floating constants denoted as $\beta_i$. The notation U$(a, b, c)$ denotes the inclusion of $c$ random points uniformly sampled from the open interval $(a, b)$ for each input variable $x_i$.}
    \label{tab:sr_benchmark}
\end{table}

\subsubsection{Analysis of traversal lengths generated in symbolic regression task.}
\begin{figure}[h!]
    \centering
    \includegraphics[width=0.7\linewidth]{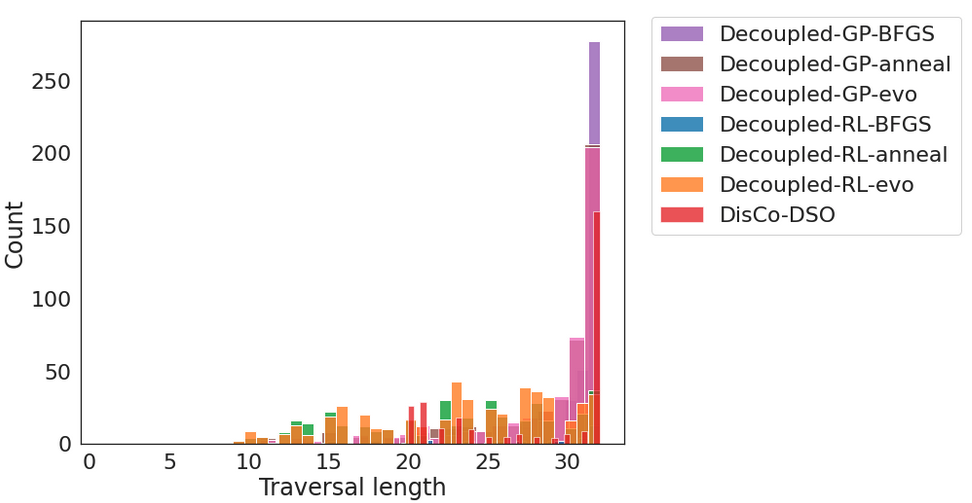}
    \caption{Histogram illustrating the lengths of traversals selected by different methods on the SR task.}
    \label{fig:sr-traversal-lengths}
\end{figure}

Figure \ref{fig:sr-traversal-lengths} shows the distribution of traversal lengths sampled by each method. These traversals are gathered from the top-performing expressions sampled, in the same way as the traversals used to generate Figure \ref{fig:sr_results}. Notice the extreme density placed at the maximum length (32) by the Decoupled-GP methods. This, paired with its poor generalization capability demonstrated in Figure \ref{fig:sr_results}a, leads us to conclude that the Decoupled-GP methods overfit heavily on the training data. The Decoupled-RL methods do this to a lesser degree, as does DisCo-DSO.

\subsubsection{Analysis of the effect of architecture and size of the autoregressive model in the symbolic regression task.}
In \Cref{tab:disco-dso-models}, we provide a compilation of results obtained by using different autoregressive models of various sizes. 
Specifically, we consider a GRU and a LSTM recurrent cell with 16, 32 and 64 hidden units. 
We observe that the performance of the model is not very sensitive to the size of the model, although LSTM models tend to perform slightly better than GRU models.
In this work, we use a LSTM model with 32 hidden units for all experiments.

\begin{table}
    \begin{center}
    \begin{tabular}{c|c|c}
         Architecture & Mean R & Std R \\ \hline
         DisCo-DSO-GRU16 & 0.7377 & 0.3161 \\
         DisCo-DSO-GRU32 & 0.7092 & 0.3442 \\
         DisCo-DSO-GRU64 & 0.7236 & 0.3261 \\ 
         DisCo-DSO-LSTM16 & 0.7385 & 0.3177 \\ 
         DisCo-DSO-LSTM32 & 0.7391 & 0.3134 \\ 
         DisCo-DSO-LSTM64 & 0.7302 & 0.3182 \\
    \end{tabular}
    \end{center}
    \caption{Experiments with symbolic regression repeated using different autoregressive models of various sizes. Specifically, we consider a GRU and a LSTM recurrent cell with 16, 32 and 64 hidden units.}
    \label{tab:disco-dso-models}
\end{table}

\subsubsection{Objective gap between the best solutions found by 
\ALGNAME and baselines for symbolic regression for equation discovery task.}
In \Cref{tab:sr_results_gap} and \Cref{tab:sr_results_literature_gap}, we show the objective gap between the best solutions found by \ALGNAME and the baselines for the symbolic regression task. The gap is computed by taking the best \ALGNAME solutions and the best solutions of the baselines.

\begin{table}[h]
    \begin{center}
    \begin{tabular}{c|c|c}
        Baseline & Dim $\leq$ 3 & All Table 3 \\
        \hline
        Decoupled-RL-BFGS & 0.0612 & 0.0645 \\
        Decoupled-RL-evo & 0.6308 & 0.6085 \\
        Decoupled-RL-anneal & 0.5459 & 0.5609 \\
        Decoupled-GP-BFGS & 0.126 & 0.2092 \\
        Decoupled-GP-evo & 0.5644 & 0.6298 \\
        Decoupled-GP-anneal & 0.5017 & 0.5681 \\
    \end{tabular}
    \end{center}
    \caption{Objective gap between the best solutions found by \ALGNAME and the baselines for the symbolic regression task. Positive values indicate that \ALGNAME found better solutions.}
    \label{tab:sr_results_gap}
\end{table}

\begin{table}[h]
    \begin{center}
    \begin{tabular}{c|c|c}
        Literature & Dim $\leq$ 3 & All Table 3 \\
        \hline
        Kamienny et al. (2022) & 0.0564 & 0.1346 \\
        Biggio et al. (2021) & -0.0226 & N/A \\
    \end{tabular}
    \end{center}
    \caption{Objective gap between the best solutions found by \ALGNAME and the baselines for the symbolic regression task. Positive values indicate that \ALGNAME found better solutions.}
    \label{tab:sr_results_literature_gap}
\end{table}


\section{Hyperparameters}

In \Cref{tab:dso-hyperparameters}, we provide the common hyperparameters used for the RL-based generative 
methods (\ALGNAME and Decoupled-RL).
The hyperparameters for the GP-based method are provided in \Cref{tab:gp-hyperparameters}.
\ALGNAME's specific hyperparameters, linked to modeling of the distribution $\mathcal{D}$,
are provided in \Cref{tab:disco-dso-hyperparameters}.
For the DT policies for reinforcement learning task,
the parameter $N$ (number of episodes to average over to compute a single reward $R(\tau)$) is set to 100.


\begin{table}[b!]
    \begin{center}
    \begin{tabular}{c|c}
        Parameter & Value \\ \hline
        Optimizer & Adam \citep{kingma2017adam} \\
        Number of layers & 1 \\
        Number of hidden units & 32 \\
        RNN type & LSTM \citep{hochreiter1997long} \\
        Learning rate ($\alpha$) & 0.001  \\
        Entropy coefficient ($\lambda_\mathcal{H}$) & 0.01  \\
        Moving average coefficient $(\beta)$ & 0.5  \\        
        Risk factor $(\epsilon)$ & 0.2 \\
    \end{tabular}
    \end{center}
    \caption{\ALGNAME and Decoupled-RL hyperparameters}
    \label{tab:dso-hyperparameters}
\end{table}

\begin{table}[b!]
    \begin{center}
    \begin{tabular}{c|c}
        Parameter & Value \\ \hline
        Population size & 1,000 \\
        Generations & 1,000 \\
        Fitness function & NRMSE \\
        Initialization method & Full \\
        Selection type & Tournament \\
        Tournament size $(k)$ & 5 \\
        Crossover probability & 0.5 \\
        Mutation probability & 0.5 \\
        Minimum subtree depth $(d_\textrm{min})$ & 0 \\
        Maximum subtree depth $(d_\textrm{max})$ & 2 \\
    \end{tabular}
    \end{center}
    \caption{Decoupled-GP hyperparameters}
    \label{tab:gp-hyperparameters}
\end{table}

\begin{table}[b!]
    \begin{center}
    \begin{tabular}{c|c}
        Parameter & Value \\ \hline
        Parameter shift & 0.0 \\
        Parameter generating distribution scale $(\sigma)$ & 0.5 \\
        Learn parameter generating distribution scale & False \\
        Parameter generating distribution type & Normal \\
    \end{tabular}
    \end{center}
    \caption{DisCo-DSO specific hyperparameters}
    \label{tab:disco-dso-hyperparameters}
\end{table}
\label{app:hyper}



\section{Performance Analysis}

Our numerical experiments were conducted using 24 cores in parallel of an Intel Xeon E5-2695 v2 machine with 128 GB per node. The experiments were implemented in Python using TensorFlow.

For Decision Tree Policies for Reinforcement Learning, every single objective function evaluation requires running $N=100$ episodes of a reinforcement learning environment. In addition, the environment is reset after every episode, and the policy is evaluated on a new environment seed. This means that every single objective function evaluation requires running $N=100$ episodes on a new environment seed. Each episode requires running a decision tree policy on a full dynamical simulation of the environment. This task is computationally expensive, and the time per function evaluation is shown in \Cref{tab:time_per_evaluation}.

\begin{table}[h]
    \begin{center}
    \begin{tabular}{c|c}
        Task & Time per function evaluation \\
        \hline
        CartPole-v1 & 0.89 s \\
        MountainCar-v0 & 1.12 s \\
        LunarLander-v2 & 1.66 s \\
        Acrobot-v1 & 5.41 s \\
    \end{tabular}
    \end{center}
    \caption{Time per function evaluation for the Decision Tree Policies for Reinforcement Learning task.}
    \label{tab:time_per_evaluation}
\end{table}

To quantify improvements in terms of computational time of \ALGNAME over the decoupled baselines, 
we provide in \Cref{tab:performance_analysis}
the average \textit{computational efficiency} for all the environments in the Decision Tree Policies for Reinforcement Learning task. We define \textit{computational efficiency} as the ratio between the final objective value and the total time required to reach that value. We compare \ALGNAME with the decoupled baselines.

\begin{table}[h]
    \begin{center}
    \begin{tabular}{c|c|c|c|c}
        Algorithm & CartPole-v1 & MountainCar-v0 & LunarLander-v2 & Acrobot-v1 \\
        \hline
        DisCo-DSO & \textbf{485.94} & \textbf{-85.87} & \textbf{-8.07} & \textbf{-11.91} \\
        Decoupled-RL-BFGS & 233.25 & -91.88 & -47.03 & -12.58 \\
        Decoupled-RL-anneal & 485.94 & -91.34 & -49.53 & -12.63 \\
        Decoupled-RL-evo & 485.94 & -87.87 & -50.05 & -12.95 \\
        Decoupled-GP-BFGS & 228.39 & -91.80 & -41.82 & -12.66 \\
        Decoupled-GP-anneal & 485.94 & -90.57 & -30.57 & -12.63 \\
        Decoupled-GP-evo & 485.94 & -88.03 & -36.51 & -13.17 \\
    \end{tabular}
    \end{center}
    \caption{Computational efficiency for the Decision Tree Policies for Reinforcement Learning task.}
    \label{tab:performance_analysis}
\end{table}

We can see that, as the scale of the problem increase (see \Cref{tab:time_per_evaluation}), 
\ALGNAME
shows a significant improvement in terms of computational time compared to the decoupled baselines.

\section{Prefix-dependent positional constraints}
\label{app:constraints}

The autoregressive sampling used by \ALGNAME 
allows for the incorporation of task-dependent constraints.
These constraints are applied \textit{in situ}, i.e., during the sampling process.
These ideas have been used by several works using similar autoregressive sampling procedures 
\citep{popova2019molecularrnn,petersen2021deep,petersen2021incorporating,landajuela2021improving,mundhenk2021symbolic,kim2021distilling}.
In this work, we include three novel constraints that are specific to the decision tree generation task, 
one on the continuous parameters and two on the discrete tokens. 

\subsection{Constraints for decision tree generation}
\label{app:priors}

\subsubsection{Constraint on parameter range.}
By using a truncated normal distribution, 
upper/lower bounds are imposed on the parameters of decision trees (i.e., $\beta_n$ in the Boolean expression tokens $x_n < \beta_n$) to 
prevent meaningless internal nodes from being sampled. 
In \Cref{alg:decision-tree-parameter-bounds}, we provide the 
detailed procedure for determining the upper/lower bounds at each position 
of the traversal. The resolution $h>0$ is a hyperparameter that controls the 
distance between the parameters $\beta_p$ at the parent node and the corresponding
 bounds at the children nodes. This guarantees that the sampled decision trees 
 must have finite depth if the environment-enforced bounds on the features of 
 the optimization problem are finite. Moreover, it also prevents the upper/lower 
 bounds from being too close, which can lead to numerical instability in the truncated normal distribution.

\begin{figure}[h!]
\begin{algorithm}[H]
\small
\caption{Finding bounds for parameters in decision trees}
\label{alg:decision-tree-parameter-bounds}
\begin{algorithmic}
\STATE \textbf{input:} parent token $l_{p}(\beta_p)$, bounds for the parameters of the parent token $\beta_p^{\max}$, $\beta_p^{\min}$
\STATE \textbf{Parameters:} Resolution $h>0$
\STATE \textbf{output:} bounds for the parameters of the next token $\beta^{\max}_i$, $\beta^{\min}_i$
\STATE $\beta_i^{\max}, \beta_i^{\min} \leftarrow \beta_p^{\max}, \beta_p^{\min}$
\RCOMMENT{Inherit parameter bounds from parent}
\IF {the next token is a right child of $l_p(\beta_p)$}
    \STATE $(\beta_i^{\min})_{l_p} \leftarrow \beta_p + h$
    \COMMENT{Adjust the lower bound corresponding to $l_p$}
    \IF {$(\beta_i^{\max})_{l_p} - (\beta_i^{\min})_{l_p} < h$ }
        \STATE $(\beta_i^{\min})_{l_p} \leftarrow (\beta_i^{\max})_{l_p} - h/2$
        \RCOMMENT{Maintain a minimal distance between bounds}
    \ENDIF
\ELSE
    \STATE $(\beta_i^{\max})_{l_p} \leftarrow \beta_p - h$
    \RCOMMENT{Adjust the upper bound corresponding to $l_p$}
    \IF {$(\beta_i^{\max})_{l_p} - (\beta_i^{\min})_{l_p} < h$ }
        \STATE $(\beta_i^{\max})_{l_p} \leftarrow (\beta_i^{\min})_{l_p} + h/2$
        \RCOMMENT{Maintain a minimal distance between bounds}
    \ENDIF
\ENDIF
\STATE \textbf{Return:} $\beta_i^{\max}, \beta_i^{\min}$

\end{algorithmic}
\end{algorithm}
\end{figure}

\subsubsection{Constraint on Boolean expression tokens.}
Depending on the values of the parameter bounds, we also impose constraints on the discrete tokens $x_n < \beta_n$. Specifically, when the upper/lower bounds $\beta_n^{\max}$ and $\beta_n^{\min}$ for the parameter of the $n$-th Boolean expression token $x_n < \beta_n$ are too close, oftentimes there is not much value to split the $n$-th feature space further. Therefore, if $\beta_n^{\max} - \beta_n^{\min} < h$, where $h$ is the resolution hyperparameter in Algorithm~\ref{alg:decision-tree-parameter-bounds}, then $x_n < \beta_n$ are constrained from being sampled.

\subsubsection{Constraint on discrete action tokens.}
If the left child and right child of a Boolean expression token $x_n < \beta_n$ are the same discrete action token $a_j$, the subtree will just be equivalent to a single leaf node containing $a_j$. We add a constraint that if the left child of $x_n < \beta_n$ is $a_j$, then the right child cannot be $a_j$.

\subsection{Constraints for equation generation}
\label{app:eq_constraints}

\subsubsection{Trigonometry constraint.}
The design space $\mathbb{T}$ in symbolic regression 
does not include expressions involving nested trigonometric functions,
such as $\sin(\cos(\sin(\dots)))$, since such expressions are not found in physical or engineering domains.
Following \citet{petersen2021deep}, 
we use a constraint to prevent the sampling of nested trigonometric functions.
For example, given the partial traversal 
$\tau = \langle +, x_1, \cos, \div, 1 \rangle$,
$\sin$ and $\cos$ 
are constrained because they would be descendants of $\cos$.

\subsubsection{Length constraint.}
For symbolic regression, we constrain the length of the traversal to prevent the generation of overly complex expressions.
We follow \citet{petersen2021deep} and constrain the length of the traversal to be no less than 4 and no more than 32.
The requirement for a minimum length is enforced by limiting terminal tokens when their selection would prematurely conclude the traversal before reaching the specified minimum length. For instance, in the case of the partial traversal $\tau = \langle \sin \rangle$, terminal tokens are restricted because opting for one would terminate the traversal with a length of 2.

On the other hand, the restriction on maximum length is applied by constraining unary and/or binary tokens when their selection, followed by the choice of only terminal tokens, would lead to a traversal surpassing the prescribed maximum length.


\end{document}